\documentclass[10pt,twocolumn,letterpaper]{article}

\usepackage[pagenumbers]{cvpr}      % To produce the REVIEW version
\usepackage[dvipsnames]{xcolor}

\definecolor{cvprblue}{rgb}{0.21,0.49,0.74}
\usepackage[pagebackref,breaklinks,colorlinks,citecolor=cvprblue]{hyperref}
\usepackage{multirow}
\usepackage{multicol}
\usepackage[ruled]{algorithm2e}
\usepackage{subcaption}
\usepackage[accsupp]{axessibility}

\DeclareMathOperator*{\argmax}{arg\,max}

\def\paperID{2426} % *** Enter the Paper ID here
\def\confName{CVPR}
\def\confYear{2024}

\title{Efficient Dataset Distillation via Minimax Diffusion}

\author{Jianyang Gu$^{1}\quad$Saeed Vahidian$^{2}\quad$Vyacheslav Kungurtsev$^{3}$\\
Haonan Wang$^{4}\quad$Wei Jiang$^{1}\thanks{Corresponding author}\quad$Yang You$^{4}\quad$Yiran Chen$^{2}\quad$\\
$^{1}$Zhejiang University$\quad^{2}$Duke University$\quad^{3}$Czech Technical University
\thanks{This work has received funding from the European Union’s Horizon Europe research and innovation program under grant agreement No. 101084642.}
\\
$^{4}$National University of Singapore\\
{\tt\small \{gu\_jianyang, jiangwei\_zju\}@zju.edu.cn}
}

\begin{document}
\maketitle

\begin{abstract}
    Dataset distillation reduces the storage and computational consumption of training a network by generating a small surrogate dataset that encapsulates rich information of the original large-scale one. However, previous distillation methods heavily rely on the sample-wise iterative optimization scheme. As the images-per-class (IPC) setting or image resolution grows larger, the necessary computation will demand overwhelming time and resources. In this work, we intend to incorporate generative diffusion techniques for computing the surrogate dataset. Observing that key factors for constructing an effective surrogate dataset are representativeness and diversity, we design additional minimax criteria in the generative training to enhance these facets for the generated images of diffusion models. We present a theoretical model of the process as hierarchical diffusion control demonstrating the flexibility of the diffusion process to target these criteria without jeopardizing the faithfulness of the sample to the desired distribution. The proposed method achieves state-of-the-art validation performance while demanding much less computational resources. Under the 100-IPC setting on ImageWoof, our method requires less than one-twentieth the distillation time of previous methods, yet yields even better performance. Source code and generated data are available in \href{https://github.com/vimar-gu/MinimaxDiffusion}{https://github.com/vimar-gu/MinimaxDiffusion}.
    % This work opens up new possibilities for the broad application of dataset distillation for personalized data with affordable resource requirements. 
\end{abstract}

\section{Introduction}
% background
Data, as a necessary resource for deep learning, has concurrently promoted algorithmic advancements while imposing challenges on researchers due to heavy demands on storage and computational resources~\cite{dosovitskiyImageWorth16x162022,he2016deep,wang2018dataset,dengImageNetLargescaleHierarchical2009}. 
Confronted with the conflict between the requirement for high-precision models and overwhelming resource demands, dataset distillation is proposed to condense the rich information of a large-scale dataset into a small surrogate one~\cite{wang2018dataset,zhao2020dataset,kim2022dataset,cui2022dc}. 
Such a surrogate dataset is expected to achieve training performance comparable to that attained with the original one. 

\begin{figure}[t]
    \centering
    \includegraphics[width=0.95\columnwidth]{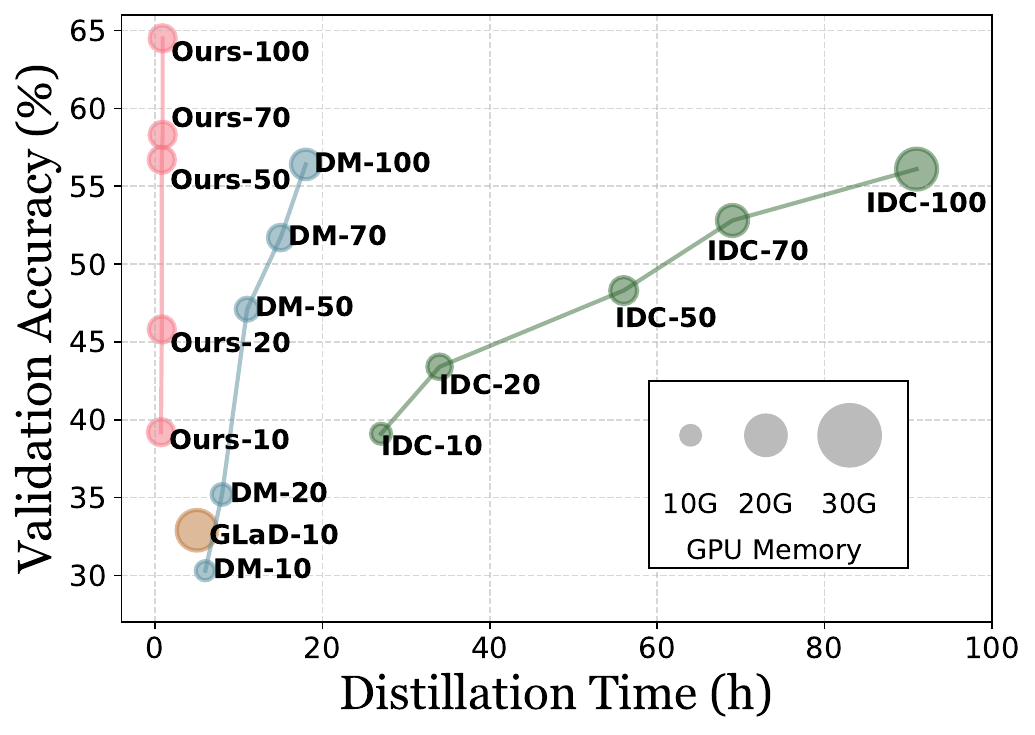}
    \caption{The validation accuracy and distillation time of different methods on ImageWoof~\cite{imagenette}, with a number following each method denoting the Image-Per-Class (IPC) setting. Previous methods are restricted by the heavier running time and memory consumption as IPC grows larger. In comparison, our proposed method notably reduces the demanding computational resources and also achieves state-of-the-art validation performance. }
    \label{fig:intro}
\end{figure}

\begin{figure}[t]
    \centering
    \includegraphics[width=0.98\columnwidth]{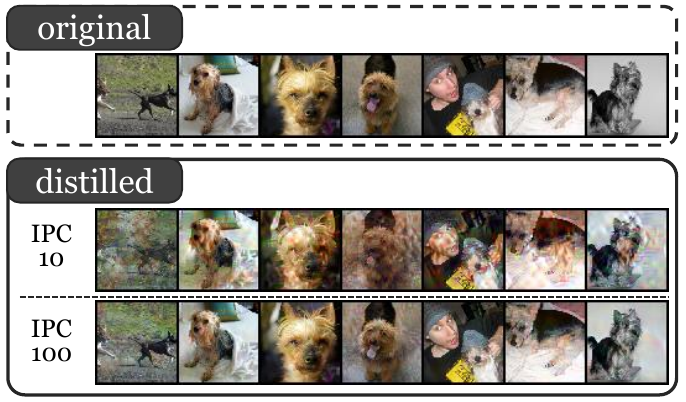}
    \caption{Sample images distilled by the pixel-level sample-wise optimization method DM~\cite{zhao2023dataset} on ImageWoof. As the parameter space increases along with the Image-Per-Class (IPC) setting, with the same initialization, the appearance disparity between original and distilled images is smaller. }
    \label{fig:dm_images}
\end{figure}

% previous problems
Previous dataset distillation methods mostly engage in iterative optimization on fixed-number samples at the pixel level~\cite{zhao2020dataset,zhao2021dataset,kim2022dataset,liu2023dream,nguyen2021dataset,nguyen2021datasetkrr,vahidian2024group} or embedding level~\cite{cazenavette2023generalizing,zhao2022synthesizing}. 
However, the sample-wise iterative optimization scheme suffers from problems of two perspectives. 
(1) The parameter space of optimization is positively correlated with the size of the target surrogate dataset and the image resolution~\cite{zhao2020dataset,cazenavette2022dataset}. 
Consequently, substantial time and computational resources are required for distilling larger datasets. 
As shown in~\cref{fig:intro}, IDC-1~\cite{kim2022dataset} takes over 90 hours to distill a 100-image-per-class (IPC) set from ImageWoof~\cite{imagenette}, while training on ImageWoof itself only requires a matter of hours. 
(2) The expanded parameter space also increases the optimization complexity.
% , resulting in marginal pixel modification as well as performance improvement. 
As shown in~\cref{fig:dm_images}, while distillation yields significant information condensation under small IPC settings, the pixel modification diminishes when distilling larger-IPC datasets. 
The reduced disparity also leads to smaller performance gain compared with original images, with instances where the distilled set even performs worse. 
Especially when distilling data of fine-grained classes, the sample-wise optimization scheme fails to provide adequate discriminative information. 
These constraints severely hinder individual researchers from distilling personalized data. 
A more practical training scheme is urgently needed to facilitate the broader application of dataset distillation. 

% proposal
In this work, we explore the possibility of incorporating generative diffusion techniques~\cite{ho2020DenoisingDiffusionProbabilistic,kingmaVariationalDiffusionModels2021,nicholImprovedDenoisingDiffusion2021} to efficiently compute effective surrogate datasets. 
We first conduct empirical analysis on the suitability of data generated by raw diffusion models for training networks. 
Based on the observations, we conclude that constructing an effective surrogate dataset hinges on two key factors: representativeness and diversity. 
Accordingly, we design extra minimax criteria for the generative training to enhance the capability of generating more effective surrogate datasets without explicit prompt designs. 
The minimax criteria involve two aspects: enforcing the generated sample to be close to the farthest real sample, while being far away from the most similar generated one. 
We provide theoretical analysis to support that the proposed minimax scheme aims to solve a well defined problem with all the criteria, including the generative accuracy and the minimax criteria, can be targeted simultaneously without detriment to the others. 
% Besides, we incorporate a variance-based sampling adjustment strategy to flexibly control the balance between the representativeness and diversity of data generation. 

% effects
Compared with the astronomical training time consumption of the sample-wise iterative optimization schemes, the proposed method takes less than 1 hour to distill a 100-IPC surrogate dataset for a 10-class ImageNet subset, including the fine-tuning and image generation processes. 
Remarkably, the GPU consumption remains consistent across all IPC settings. 
Furthermore, the distilled surrogate dataset attains superior validation performance compared with other state-of-the-art methods. 
Especially on the challenging fine-grained ImageWoof subset, the proposed method outperforms the second-best DD method by 5.5\% and 8.1\% under the IPC settings of 70 and 100, respectively. 
The source code is provided in the supplementary material. 
% Besides, the proposed method also generates better results on the fine-grained ImageWoof subset, demonstrating the probability of broader applications. 
% For example, under the settings of 70 and 100 IPC, our method surpasses the second best results by 2.1\% and 0.6\%, respectively. 

% contributions
The contributions of this work are summarized into:
\begin{itemize}
    \item We analyze the data generated by diffusion models, and emphasize the importance of representativeness and diversity for constructing effective surrogate datasets. 
    \item We propose a novel dataset distillation scheme based on extra minimax criteria for diffusion models targeting the representativeness and diversity of generated data.
    \item We theoretically justify the proposed minimax criteria as enforceable without trade-offs in the generation quality of the individual data points.
    \item We conduct extensive experiments to validate that our proposed method achieves state-of-the-art performance while demanding significantly reduced training time in comparison to previous dataset distillation methods.
\end{itemize}

\begin{figure}[t]
    \centering
    \includegraphics[width=\columnwidth]{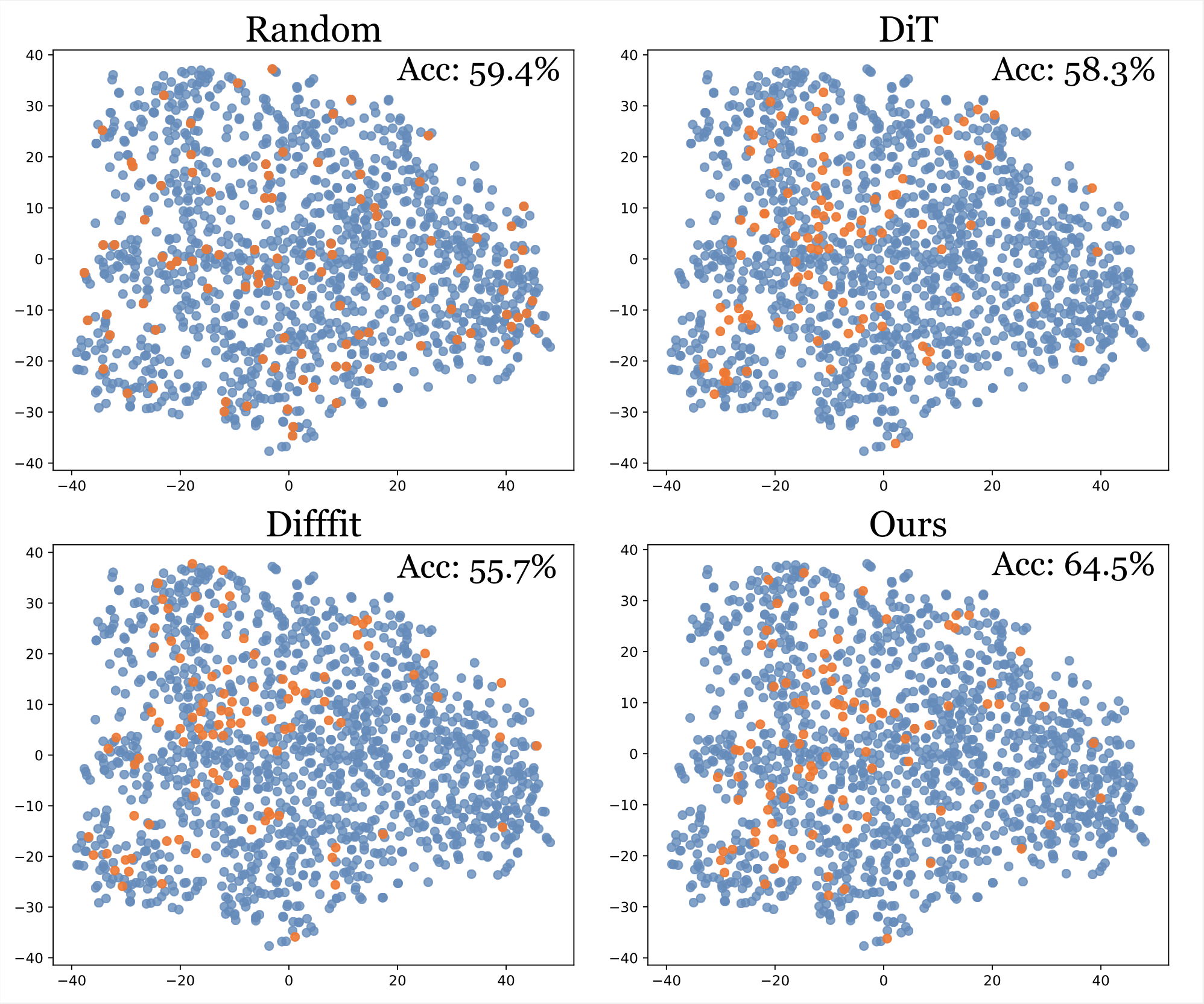}
    \caption{The feature distribution comparison of different image generation methods with the original set. The validation performance of each surrogate set is listed in the upper-right corner. }
    \label{fig:tsne-dit}
\end{figure}

\section{Method}

\subsection{Problem Definition}
The general purpose of dataset distillation is to generate a small surrogate dataset $\mathcal{S}=\{(\mathbf{x}_i,y_i)\}^{N_S}_{i=1}$ from a large-scale one $\mathcal{T}=\{(\mathbf{x}_i,y_i)\}^{N_T}_{i=1}$~\cite{wang2018dataset,zhao2020dataset}. 
Here each $\mathbf{x}_i$ denotes an image with a corresponding class label $y_i$, and $N_S \ll N_T$. 
The surrogate dataset $\mathcal{S}$ is expected to encapsulate substantial information from the original $\mathcal{T}$, such that training a model on $\mathcal{S}$ achieves performance comparable with that on $\mathcal{T}$. 
After distilling, we train network models on $\mathcal{S}$ and validate the performance on the original test set. 

\subsection{Diffusion for Distillation}
Diffusion models learn a dataset distribution by gradually adding Gaussian noise to images and reversing back. 
Taking the latent diffusion model (LDM) as an example, given a training image $\mathbf{x}$, the training process is separated into two parts. 
An encoder $E$ transforms the image into the latent space $\mathbf{z}=E(\mathbf{x})$ and a decoder $D$ reconstructs a latent code back to the image space $\hat{\mathbf{x}}=D(\mathbf{z})$. 
The forward noising process gradually adds noise $\epsilon \sim \mathcal{N}(\mathbf{0},\mathbf{I})$ to the original latent code $\mathbf{z}_0$: $\mathbf{z}_{t}=\sqrt{\bar{\alpha}_t}\mathbf{z}_0+\sqrt{1-\bar{\alpha}_t}\epsilon$, where $\bar{\alpha}_t$ is a hyper-parameter. 
Provided with a conditioning vector $\mathbf{c}$ encoded with class labels, the diffusion models are trained by the squared error between the predicted noise $\epsilon_\theta(\mathbf{z}_t,\mathbf{c})$ and the ground truth $\epsilon$:
\begin{equation}\label{eq:simple}
    \mathcal{L}_{simple}=||\epsilon_\theta(\mathbf{z}_t,\mathbf{c})-\epsilon||^2_2,
\end{equation}
where $\epsilon_\theta$ is a noise prediction network parameterized by $\theta$. 
Diffusion models are proven to generate images of higher quality compared with GANs~\cite{dhariwalDiffusionModelsBeat2021}. 
There are also some Parameter Efficient Fine-Tuning (PEFT) methods updating a small number of model parameters in order for the model to be better adapted to specific data domains~\cite{ruiz2023dreambooth,xie2023difffit}. 
We adopt DiT~\cite{peebles2023scalable} as the baseline and Difffit~\cite{xie2023difffit} as the naive fine-tuning method for image generation. 
The generated images are compared with the original data from the perspective of embedding distribution in~\cref{fig:tsne-dit}. 

The samples of random selection and pre-trained diffusion models present two extreme ends of the distribution. 
Random selection faithfully reflects the original distribution, yet fails to emphasize some high-density regions. 
In contrast, diffusion models are over-fitted to those dense areas, leaving a large part of the original distribution uncovered. 
We attribute these two distributions to two properties, respectively. 
The randomly selected data holds extraordinary \textit{diversity}, and the diffusion-generated data shows \textit{representativeness} to the original distribution. 
We claim that both properties are essential for constructing an effective surrogate dataset. 
By naive fine-tuning, Difffit better captures the representative regions but leaves more regions uncovered. 
To this end, we propose extra minimax criteria for the diffusion model to enhance both of the properties. 

\subsection{Minimax Diffusion Criteria}
Based on the observation that \textit{representativeness} and \textit{diversity} are two key factors to construct an effective surrogate dataset, we accordingly design extra minimax criteria to enhance these two essential properties for the diffusion model. 
% The minimax optimization pipeline is shown in Fig.~\ref{fig:framework}.

\paragraph{Representativeness}
It is essential for the small surrogate dataset to sufficiently represent the original data. 
A naive approach to improve the representativeness is aligning the embedding distribution between synthetic and real samples:
\begin{equation}
\label{eq:naive_repre}
    \mathcal{L}_{r}=\argmax_{\theta}\sigma\left(\hat{\mathbf{z}}_\theta(\mathbf{z}_t,\mathbf{c}),\frac{1}{N_B}\sum_{i=0}^{N_B}\mathbf{z}_i\right),
\end{equation}
where $\sigma(\cdot,\cdot)$ is the cosine similarity, $\hat{\mathbf{z}}_\theta(\mathbf{z}_t,\mathbf{c})$ is the predicted original embedding by subtracting the noise from the noisy embedding $\hat{\mathbf{z}}_\theta(\mathbf{z}_t,\mathbf{c})=\mathbf{z}_t-\epsilon_\theta(\mathbf{z}_t,\mathbf{c})$, and $N_B$ is the size of the sampled real sample mini-batch. 
However, the naive alignment tends to draw the predicted embedding towards the center of the real distribution, which severely limits the diversity. 
Therefore, we propose to maintain an auxiliary memory $\mathcal{M}=\{\mathbf{z}_m\}^{N_M}_{m=1}$ to store the real samples utilized in adjacent iterations, and design a minimax optimization objective as:
\begin{equation}
\label{eq:minimax_repre}
    \mathcal{L}_{r}=\arg\max_{\theta}\min_{m\in [N_M]}\sigma\left(\hat{\mathbf{z}}_\theta(\mathbf{z}_t,\mathbf{c}),\mathbf{z}_m\right).
\end{equation}
By pulling close the least similar sample pairs, the diffusion model is encouraged to generate images that better cover the original distribution. 
It is notable that the diffusion training objective $\mathcal{L}_{simple}$ itself encourages the generated images to resemble the original ones. 
Thus, the minimax criterion allows the preservation of diversity to the maximum extent. 

\begin{algorithm}[t]
\DontPrintSemicolon
\caption{Minimax Diffusion Fine-tuning}
\label{alg:minimax}
\small
\KwIn{initialized model parameter $\theta$, original dataset $\mathcal{T}=\{(\mathbf{x},y)\}$, encoder $E$, class encoder $E_c$, time step $t$, variance schedule $\bar{\alpha}_t$, real embedding memory $\mathcal{M}$, predicted embedding memory $\mathcal{D}$}
\KwOut{optimized model parameter $\theta^*$}
\SetKwBlock{Begin}{function}{end function}{
\For{each step}{
    Obtain the original embedding: $\mathbf{z}_0=E(\mathbf{x})$\;
    Obtain the class embedding: $\mathbf{c}=E_c(y)$\;
    Sample random noise: $\epsilon \sim \mathcal{N}(\mathbf{0},\mathbf{I})$\;
    Add noise to the embedding: $\mathbf{z}_{t}=\sqrt{\bar{\alpha}_t}\mathbf{z}_0+\sqrt{1-\bar{\alpha}_t}\epsilon$\;
    Predict the noise $\epsilon_\theta(\mathbf{z}_t,\mathbf{c})$ and recovered embedding $\hat{\mathbf{z}}_\theta(\mathbf{z}_t,\mathbf{c})=\mathbf{z}_t-\epsilon_\theta(\mathbf{z}_t,\mathbf{c})$\;
    Update the model parameter with~\cref{eq:total}\;
    Enqueue the real embedding: $\mathcal{M}_r\leftarrow\mathbf{z}_0$\;
    Enqueue the predicted embedding: $\mathcal{M}_d\leftarrow\hat{\mathbf{z}}_\theta(\mathbf{z}_t,\mathbf{c})$\;
}
}
\end{algorithm}

\paragraph{Diversity}
Although the pre-trained diffusion models already achieve satisfactory generation quality, the remaining defect is limited diversity compared with the original data, as shown in~\cref{fig:tsne-dit}.
We expect the data generated by the diffusion model can accurately reflect the original distribution, while simultaneously being different from each other. 
Hence, we maintain another auxiliary memory $\mathcal{D}=\{\mathbf{z}_d\}^{N_D}_{d=1}$ for the predicted embeddings of adjacent iterations and design another minimax objective to explicitly enhance the sample diversity as:
\begin{equation}\label{eq:lossdiversity}
    \mathcal{L}_{d}=\arg\min_\theta\max_{d\in [N_D]}\sigma\left(\hat{\mathbf{z}}_\theta(\mathbf{z}_t,\mathbf{c}),\mathbf{z}_d\right).
\end{equation}
The diversity term has an opposite optimization target compared with the representativeness term, where the predicted embedding is pushed away from the most similar one stored in the memory bank. 
Although diversity is essential for an effective surrogate set, too much of it will cause the generated data to lose representativeness. 
The proposed minimax optimization enhances the diversity in a gentle way, with less influence on the class-related features. 

Combining all the components, we summarize the training process in~\cref{alg:minimax}. 
The complete training objective can be formulated as:
\begin{equation}
\label{eq:total}
    \mathcal{L}=\mathcal{L}_{simple}+\lambda_r\mathcal{L}_r+\lambda_d\mathcal{L}_d,
\end{equation}
where $\lambda_r$ and $\lambda_d$ are weighting hyper-parameters.
% and note that we minimize the negative to aim for the maximum for $\mathcal{L}_r$. 

\section{Theoretical Analysis}
Assume that $\mu$ is the real distribution of the latent variables $\mathbf{z}$ associated with the target dataset $\mathcal{T}$. We rewrite the optimization problem presented in~\cref{eq:total} in a modified form:
\begin{equation}\label{eq:moo}
\small
\begin{array}{l}
    \min\limits_{\{\theta^{(i)}\}_{i\in[N_D]}} \,\lambda_d 
    \max\limits_{i,j=1,..,N_D} \sigma\left(\hat{\mathbf{z}}(\theta^{(i)}),\hat{\mathbf{z}}(\theta^{(j)})\right)\\ \quad + \sum\limits_{i=1}^{N_D}\left\{-\lambda_r Q_{\tilde{q},w\sim \mu}\left[\sigma\left(\hat{\mathbf{z}}(\theta^{(i)}),w\right)\right]+  \|\hat{\mathbf{z}}(\theta^{(i)})-\mathbf{z}^{(i)}_0\|^2\right\},
    \end{array}
\end{equation}
where $Q_{\tilde{q}}[\cdot]$ denotes the quantile function with $\tilde{q}$ as the quantile percentage. 
Note that here we consider a theoretical idealized variant of our algorithm wherein we perform simultaneous generation of all the embeddings $\{\hat{\mathbf{z}}(\theta^{(i)})\}$, rather than sample by sample.
Hence the objectives turn to the sum of pairwise similarities rather than the form in~\cref{eq:lossdiversity}.
% The enforcement of reconstruction is given by the last term in the summand.
And we minimize the negative to aim for maximal representativeness, as in~\cref{eq:total}. 

% Hence, the objectives turn to the sum of pairwise similarities compared to the similarities of one with respect to the others generated so far as~\cref{eq:lossdiversity}.
 
It can be considered as a scalarized solution to a multi-objective optimization problem, wherein multiple criteria are weighed (see, \eg~\cite{eichfelder2009scalarizations}). This perspective aligns with a Pareto front with trade-offs. It means that one objective decreasing will by necessity result in another increasing. 

However, consider that any solution to the following tri-level optimization problem is also a solution for~\cref{eq:moo}:
\begin{equation}\label{eq:finitecontrol}
\small
    \begin{array}{rl}
\min\limits_{\{\theta^{(i)}\}_{i\in[N_D]}} & \max\limits_{i,j=1,..,N_D} \sigma\left(\hat{\mathbf{z}}(\theta^{(i)}),\hat{\mathbf{z}}(\theta^{(j)})\right) \\
\text{subj. to }& \{\theta^{(i)}\} \in \arg\max \sum\limits_{i=1}^{N_D}Q_{\tilde{q},w\sim \mu}\left[\sigma\left(\hat{\mathbf{z}}(\theta^{(i)}),w\right)\right] \\ \text{subj. to}
& \theta^{(i)} \in \arg\min \|\hat{\mathbf{z}}(\theta)-\mathbf{z}^{(i)}_0\|^2,\,\forall i\in[N_D].
    \end{array}
\end{equation}
If a solution to~\cref{eq:finitecontrol} is discovered, either incidentally through solving~\cref{eq:moo} or by careful tuning of step sizes, the set of minimizers will be sufficiently large at both levels, with no trade-offs involved. However, can we justify the presumption that there exists a meaningful set of potential minimizers?

\paragraph{Diffusion Process Model} One popular framework for the mathematical analysis of diffusion involves analyzing the convergence and asymptotic properties of, appropriately homonymous, diffusion processes. These processes are characterized by the standard stochastic differential equation with a drift and diffusion term. 
For a time-dependent random variable $Z_t$,
\begin{equation}\label{eq:sde}
dZ_t=V(Z_t) dt+dW_t
\end{equation}
where $V$ is a drift function dependent on the current $Z_t$ and $dW_t$ is a Wiener (Brownian noise) process. This equation serves as an appropriate continuous approximation of generative diffusion, given that Brownian noise is a continuous limit of adding normal random variables. Consequently, we aim for any realization $\mathbf{z}\sim Z_t$ to have certain desired properties that reflect generative modeling with high probability. 

\begin{table*}[t]
    \centering
    \caption{Performance comparison with pre-trained diffusion models and other state-of-the-art methods on ImageWoof. All the results are reproduced by us on the 256$\times$256 resolution. The missing results are due to out-of-memory. The best results are marked as \textbf{bold}. }
    \label{tab:imagenet-10}
    \small
    \setlength{\tabcolsep}{4pt}
    \resizebox{\textwidth}{!}{
    \begin{tabular}{cl|ccccccc|c|c}
    \toprule
        IPC (Ratio) & Test Model & Random & K-Center~\cite{sener2018active} & Herding~\cite{welling2009herding} & DiT~\cite{peebles2023scalable} & DM~\cite{zhao2023dataset} & IDC-1~\cite{kim2022dataset} & GLaD~\cite{cazenavette2023generalizing} & Ours & Full \\
         \midrule
        \multirow{3}{*}{10 (0.8\%)}
         & ConvNet-6   & 24.3$_{\pm 1.1}$ & 19.4$_{\pm 0.9}$ & 26.7$_{\pm 0.5}$ & 34.2$_{\pm 1.1}$ & 26.9$_{\pm 1.2}$ & 33.3$_{\pm 1.1}$ & 33.8$_{\pm 0.9}$ & \textbf{37.0$_{\pm 1.0}$} & 86.4$_{\pm 0.2}$ \\
         & ResNetAP-10 & 29.4$_{\pm 0.8}$ & 22.1$_{\pm 0.1}$ & 32.0$_{\pm 0.3}$ & 34.7$_{\pm 0.5}$ & 30.3$_{\pm 1.2}$ & 39.1$_{\pm 0.5}$ & 32.9$_{\pm 0.9}$ & \textbf{39.2$_{\pm 1.3}$} & 87.5$_{\pm 0.5}$ \\
         & ResNet-18   & 27.7$_{\pm 0.9}$ & 21.1$_{\pm 0.4}$ & 30.2$_{\pm 1.2}$ & 34.7$_{\pm 0.4}$ & 33.4$_{\pm 0.7}$ & 37.3$_{\pm 0.2}$ & 31.7$_{\pm 0.8}$ & \textbf{37.6$_{\pm 0.9}$} & 89.3$_{\pm 1.2}$ \\
        \midrule
        \multirow{3}{*}{20 (1.6\%)}
         & ConvNet-6   & 29.1$_{\pm 0.7}$ & 21.5$_{\pm 0.8}$ & 29.5$_{\pm 0.3}$ & 36.1$_{\pm 0.8}$ & 29.9$_{\pm 1.0}$ & 35.5$_{\pm 0.8}$ & - & \textbf{37.6$_{\pm 0.2}$} & 86.4$_{\pm 0.2}$ \\
         & ResNetAP-10 & 32.7$_{\pm 0.4}$ & 25.1$_{\pm 0.7}$ & 34.9$_{\pm 0.1}$ & 41.1$_{\pm 0.8}$ & 35.2$_{\pm 0.6}$ & 43.4$_{\pm 0.3}$ & - & \textbf{45.8$_{\pm 0.5}$} & 87.5$_{\pm 0.5}$ \\
         & ResNet-18   & 29.7$_{\pm 0.5}$ & 23.6$_{\pm 0.3}$ & 32.2$_{\pm 0.6}$ & 40.5$_{\pm 0.5}$ & 29.8$_{\pm 1.7}$ & 38.6$_{\pm 0.2}$ & - & \textbf{42.5$_{\pm 0.6}$} & 89.3$_{\pm 1.2}$ \\
        \midrule
        \multirow{3}{*}{50 (3.8\%)}
         & ConvNet-6   & 41.3$_{\pm 0.6}$ & 36.5$_{\pm 1.0}$ & 40.3$_{\pm 0.7}$ & 46.5$_{\pm 0.8}$ & 44.4$_{\pm 1.0}$ & 43.9$_{\pm 1.2}$ & - & \textbf{53.9$_{\pm 0.6}$} & 86.4$_{\pm 0.2}$ \\
         & ResNetAP-10 & 47.2$_{\pm 1.3}$ & 40.6$_{\pm 0.4}$ & 49.1$_{\pm 0.7}$ & 49.3$_{\pm 0.2}$ & 47.1$_{\pm 1.1}$ & 48.3$_{\pm 1.0}$ & - & \textbf{56.3$_{\pm 1.0}$} & 87.5$_{\pm 0.5}$ \\
         & ResNet-18   & 47.9$_{\pm 1.8}$ & 39.6$_{\pm 1.0}$ & 48.3$_{\pm 1.2}$ & 50.1$_{\pm 0.5}$ & 46.2$_{\pm 0.6}$ & 48.3$_{\pm 0.8}$ & - & \textbf{57.1$_{\pm 0.6}$} & 89.3$_{\pm 1.2}$ \\
        \midrule
        \multirow{3}{*}{70 (5.4\%)}
         & ConvNet-6   & 46.3$_{\pm 0.6}$ & 38.6$_{\pm 0.7}$ & 46.2$_{\pm 0.6}$ & 50.1$_{\pm 1.2}$ & 47.5$_{\pm 0.8}$ & 48.9$_{\pm 0.7}$ & - & \textbf{55.7$_{\pm 0.9}$} & 86.4$_{\pm 0.2}$ \\
         & ResNetAP-10 & 50.8$_{\pm 0.6}$ & 45.9$_{\pm 1.5}$ & 53.4$_{\pm 1.4}$ & 54.3$_{\pm 0.9}$ & 51.7$_{\pm 0.8}$ & 52.8$_{\pm 1.8}$ & - & \textbf{58.3$_{\pm 0.2}$} & 87.5$_{\pm 0.5}$ \\
         & ResNet-18   & 52.1$_{\pm 1.0}$ & 44.6$_{\pm 1.1}$ & 49.7$_{\pm 0.8}$ & 51.5$_{\pm 1.0}$ & 51.9$_{\pm 0.8}$ & 51.1$_{\pm 1.7}$ & - & \textbf{58.8$_{\pm 0.7}$} & 89.3$_{\pm 1.2}$ \\
        \midrule
        \multirow{3}{*}{100 (7.7\%)}
         & ConvNet-6   & 52.2$_{\pm 0.4}$ & 45.1$_{\pm 0.5}$ & 54.4$_{\pm 1.1}$ & 53.4$_{\pm 0.3}$ & 55.0$_{\pm 1.3}$ & 53.2$_{\pm 0.9}$ & - & \textbf{61.1$_{\pm 0.7}$} & 86.4$_{\pm 0.2}$ \\
         & ResNetAP-10 & 59.4$_{\pm 1.0}$ & 54.8$_{\pm 0.2}$ & 61.7$_{\pm 0.9}$ & 58.3$_{\pm 0.8}$ & 56.4$_{\pm 0.8}$ & 56.1$_{\pm 0.9}$ & - & \textbf{64.5$_{\pm 0.2}$} & 87.5$_{\pm 0.5}$ \\
         & ResNet-18   & 61.5$_{\pm 1.3}$ & 50.4$_{\pm 0.4}$ & 59.3$_{\pm 0.7}$ & 58.9$_{\pm 1.3}$ & 60.2$_{\pm 1.0}$ & 58.3$_{\pm 1.2}$ & - & \textbf{65.7$_{\pm 0.4}$} & 89.3$_{\pm 1.2}$ \\
    \bottomrule
    \end{tabular}
    }
\end{table*}

The work~\cite{tzen2019theoretical} established a theoretical model utilizing concepts in the control of these diffusions, demonstrating how it can result in sampling from the distribution of a desired data set. In the supplementary material we present a description of their framework and present an argument supporting the well-defined nature of the following problem, indicating that it has non-trivial solutions.

When sampling optimally from the population dataset, we consider a stochastic control problem wherein $V$ depends also on some chosen control $u(\mathbf{z},t)$. This control aims to find the most representative samples and, among the possible collection of such samples, to obtain the most diverse one while sampling from the desired dataset $\mu$. This involves solving:
\begin{equation}\label{eq:bicontrol}
\small
    \begin{array}{rl}
\min\limits_{u(x,t)} & \max\limits_{i,j=1,..,N_D} \sigma\left(Z_1^{u,(i)},Z_1^{u,(j)}\right) \\
\text{subj. to }& u \in \arg\max \sum\limits_{i=1}^{N_D}\int_0^1\mathbb{E}_{Z^{(i)}_t} Q_{\tilde{q},w\sim \mu}\left[\sigma\left(Z^{(i)}_t,w\right)\right]ds \\ 
& Z_1\sim \mu,\\
& dZ^{u,(i)}_t = u(Z^{u,(i)}_t,t) dt+dW_t,\, t\in [0,1];\\
& Z_0= \mathbf{z}_0.
    \end{array}
\end{equation}
This problem poses a bi-level stochastic control challenge where employing a layered dynamic programming is far from tractable. Additionally, a multi-stage stochastic programming approximation would also be infeasible given the scale of the datasets involved. Instead, we opt for parameterization with a neural network, forego exact sampling, discretize the problem, redefine the criteria to be time independent, and seek to solve an approximate solution for the tri-level optimization problem~\cref{eq:finitecontrol}.

In the supplementary material we provide a rationale for the meaningfulness of the problem in~\cref{eq:bicontrol} based on the model of generative diffusion~\cite{tzen2019theoretical}.
Specifically, we argue that the set of controls that leads to the desired final distribution and the set of minimizers, is sufficiently large for a low value of the objective at the top layer.

\section{Experiments}

\subsection{Implementation Details}
For the diffusion model, we adopt pre-trained DiT~\cite{peebles2023scalable} as the baseline and conduct PEFT with Difffit~\cite{xie2023difffit}. 
$\lambda_r$ and $\lambda_d$ are set as 0.002 and 0.008 for~\cref{eq:total}, respectively. 
The image size for the diffusion fine-tuning and sample generation is set as 256$\times$256. 
The fine-tuning mini-batch size is set as 8, and the fine-tuning lasts 8 epochs. 
The learning rate is set as 1e-3 for an AdamW optimizer. 
After fine-tuning, the images are generated by 50 denoising steps on a pre-defined number of random noise, according to the IPC setting. 
All the experiments are conducted on a single RTX 4090 GPU. 

\subsection{Datasets and Evaluation Metric}
For practical applicability, the experiments are exclusively conducted on full-sized ImageNet~\cite{dengImageNetLargescaleHierarchical2009} subsets in this work. 
The selected subsets include ImageWoof, ImageNette~\cite{imagenette} and the 10-class split adopted in~\cite{kim2022dataset,tian2020contrastive}, denoted as ImageIDC afterward. 
ImageWoof is a challenging subset, containing only classes of dog breeds, while ImageNette and ImageIDC contain classes with less similarity, and hence are easier to discriminate. 
For evaluation, we adopt the same setting as in~\cite{kim2022dataset}. 
The surrogate dataset is trained on different model architectures, with a learning rate of 0.01, and a scheduler decaying the learning rate at 2/3 and 5/6 of the whole training iterations. 
The top-1 accuracy on the original testing set is reported to illustrate the performance. 

\subsection{Comparison with State-of-the-art Methods}
We compare our method with other state-of-the-art methods across different IPC settings and model architectures.
For a fair comparison, the results are all reproduced by us under the same evaluation protocol. 
ResNet-10~\cite{he2016deep} with average pooling is adopted for matching the feature distribution (DM~\cite{zhao2023dataset}, GLaD~\cite{cazenavette2023generalizing}) and training gradients (IDC-1~\cite{kim2022dataset}). 
% The 10-class split follows the setting in~\cite{kim2022dataset}. 
DM is implemented on IDC-1 by only modifying the matching objective from training gradients to feature distribution, such that better performance is achieved. 
Each experiment is conducted 3 times, with the mean value and standard variance reported. 
Firstly, we present the validation results on the challenging ImageWoof subset~\cite{imagenette} in~\cref{tab:imagenet-10}. 

\begin{table}[t]
    \centering
    \caption{The Maximum Mean Discrepancy (MMD) between the extracted features of surrogate dataset and the original one. }
    \label{tab:kl-div}
    \small
    \setlength{\tabcolsep}{4.8pt}
    \begin{tabular}{c|ccccc}
    \toprule
        IPC & DiT~\cite{peebles2023scalable} & Difffit~\cite{xie2023difffit} & DM~\cite{zhao2023dataset} & IDC-1~\cite{kim2022dataset} & Ours \\
    \midrule
        50 & 5.4 & 5.4 & 4.8 & 6.7 & 4.0 \\
        100 & 5.5 & 5.3 & 4.0 & 6.4 & 4.3 \\
    \bottomrule
    \end{tabular}
\end{table}

With the target of distilling surrogate datasets of small IPCs (\eg, 10 and 20), the pixel-level optimization method IDC-1~\cite{kim2022dataset} demonstrates outstanding performance gain over random original images. 
However, as the IPC increases, the performance gain drastically drops. 
Especially under the 100-IPC setting, the distilled dataset even performs worse than random original images. 
This observation aligns with the empirical findings in~\cref{fig:dm_images}, where pixel-level methods struggle to optimize the expanded parameter space of large IPCs. 
The embedding-level optimization method GLaD~\cite{cazenavette2023generalizing} yields good performance under the 10-IPC setting. 
However, it requires overwhelming GPU resources for larger IPC settings, which is inapplicable for resource-restricted scenarios. 
It is also notable that on large IPCs, the coreset method Herding~\cite{welling2009herding} surpasses previous DD methods with far less computational cost. 

\begin{table}[t]
    \centering
    \small
    \caption{Performance comparison with pre-trained diffusion models and state-of-the-art methods on more ImageNet subsets. The results are obtained on ResNet-10 with average pooling. The best results are marked as \textbf{bold}. }
    \label{tab:imagenette}
    \setlength{\tabcolsep}{7.2pt}
    \begin{tabular}{l|c|cccc}
    \toprule
        & IPC & Random & DiT~\cite{peebles2023scalable} & DM~\cite{zhao2023dataset} & Ours \\
    \midrule
        \parbox[t]{2mm}{\multirow{3}{*}{\rotatebox[origin=c]{90}{Nette}}}
        & 10 & 54.2$_{\pm 1.6}$ & 59.1$_{\pm 0.7}$ & 60.8$_{\pm 0.6}$ & \textbf{62.0$_{\pm 0.2}$} \\
        & 20 & 63.5$_{\pm 0.5}$ & 64.8$_{\pm 1.2}$ & 66.5$_{\pm 1.1}$ & \textbf{66.8$_{\pm 0.4}$} \\
        & 50 & 76.1$_{\pm 1.1}$ & 73.3$_{\pm 0.9}$ & 76.2$_{\pm 0.4}$ & \textbf{76.6$_{\pm 0.2}$} \\
        \midrule
        \parbox[t]{2mm}{\multirow{3}{*}{\rotatebox[origin=c]{90}{IDC}}}
        & 10 & 48.1$_{\pm 0.8}$ & \textbf{54.1$_{\pm 0.4}$} & 52.8$_{\pm 0.5}$ & 53.1$_{\pm 0.2}$ \\
        & 20 & 52.5$_{\pm 0.9}$ & 58.9$_{\pm 0.2}$ & 58.5$_{\pm 0.4}$ & \textbf{59.0$_{\pm 0.4}$} \\
        & 50 & 68.1$_{\pm 0.7}$ & 64.3$_{\pm 0.6}$ & 69.1$_{\pm 0.8}$ & \textbf{69.6$_{\pm 0.2}$} \\
    \bottomrule
    \end{tabular}
\end{table}

The pre-trained DiT~\cite{peebles2023scalable} here serves as the baseline for generative diffusion techniques. 
Under the 50-IPC setting, DiT outperforms both random original images and IDC-1. 
However, the insufficiency of representativeness and diversity restricts its performance on smaller and larger IPCs, respectively. 
In contrast, our proposed minimax diffusion consistently provides superior performance across all the IPCs over both original images and Herding. 
Besides, the proposed method eliminates the need of specific network architectures for matching training metrics. 
Consequently, the cross-architecture generalization is significantly improved. 
Under most IPC settings, the performance gap between ConvNet-6 and ResNetAP-10 is even smaller than that of the original images. 
It validates the universality of the rich information learned by the minimax fine-tuning process. 

Furthermore, we extensively assess the Maximum Mean Discrepancy (MMD) between the embedded features of the selected/generated surrogate dataset and the original one in~\cref{tab:kl-div}. 
The features are extracted by a ResNet-10 network pre-trained on the full original dataset. 
Our method achieves the lowest discrepancy by average, where DM~\cite{zhao2023dataset} directly sets MMD as the optimization target, proving the validity of extra minimax criteria in fitting distributions. 

Moreover, we show the performance comparison on ImageNette~\cite{imagenette} and ImageIDC~\cite{kim2022dataset} in~\cref{tab:imagenette}. 
The performance trend generally aligns with that on ImageWoof. 
More specifically, on these two easier subsets, DiT quickly loses the advantage over original images as IPC increases. 
Conversely, our proposed minimax diffusion method consistently demonstrates state-of-the-art performance. 

\begin{table}[t]
    \centering
    \caption{Performance comparison on ImageNet-1K. }
    \label{tab:im-1k}
    \small
    \setlength{\tabcolsep}{6pt}
    \begin{tabular}{c|cccc}
    \toprule
        IPC & SRe$^2$L~\cite{yin2023sre2l} & RDED~\cite{sun2024diversity} & DiT & Ours \\
    \midrule
        10 & 21.3$_{\pm 0.6}$ & 42.0$_{\pm 0.1}$ & 39.6$_{\pm 0.4}$ & \textbf{44.3$_{\pm 0.5}$} \\
        50 & 46.8$_{\pm 0.2}$ & 56.5$_{\pm 0.1}$ & 52.9$_{\pm 0.6}$ & \textbf{58.6$_{\pm 0.3}$} \\
    \bottomrule
    \end{tabular}
\end{table}

\noindent
\textbf{Experiments on ImageNet-1K.}
We further conduct experiments on the full ImageNet-1K with the validation protocol of RDED~\cite{sun2024diversity} and present the results in Tab.~\ref{tab:im-1k}. 
The synthetic images are resized to 224$\times$224 for evaluation. 
The significant performance advantage over the compared works validates the scalability of the proposed method. 

\begin{figure}[t]
    \centering
    \includegraphics[width=\columnwidth]{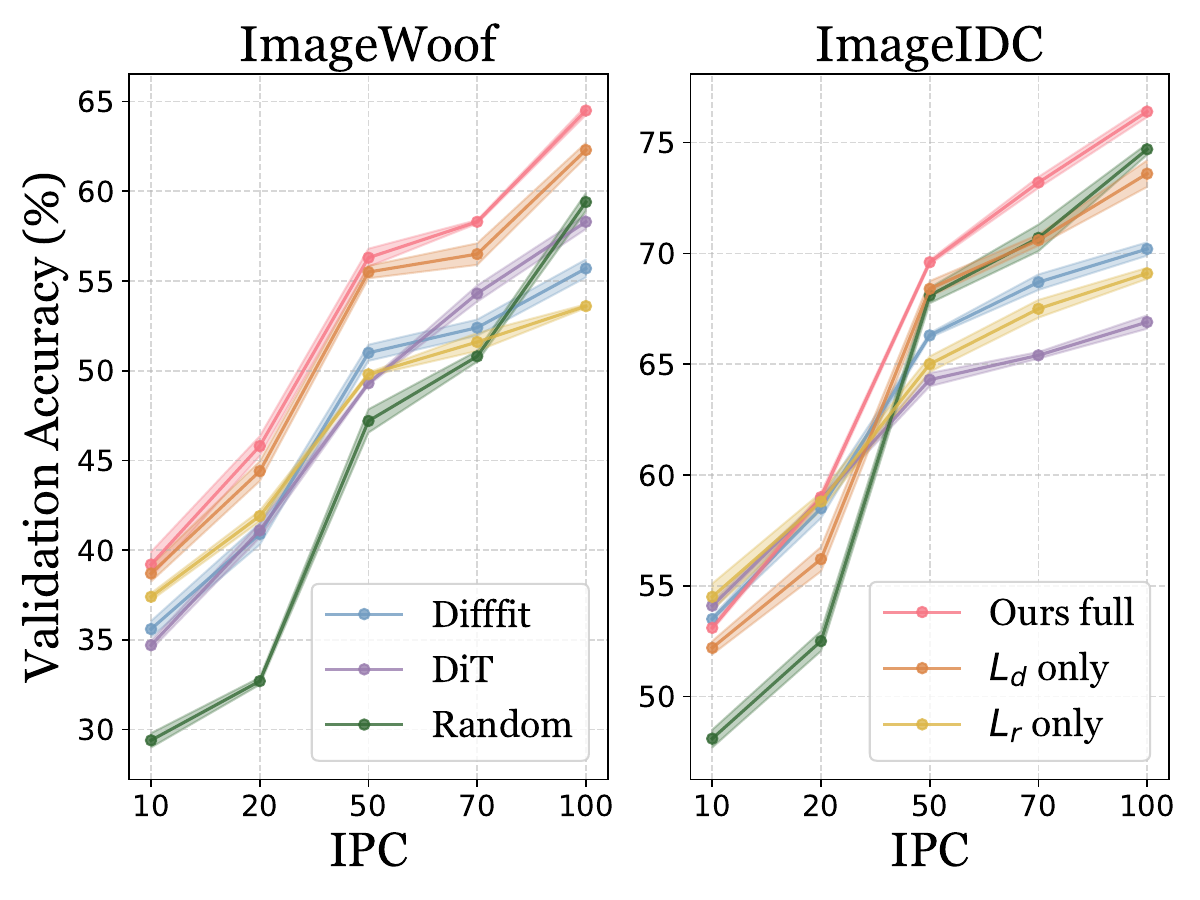}
    \caption{With the help of the minimax diffusion, the proposed method significantly enhances the representativeness and diversity of the generated images. Thereby it consistently provides superior performance compared with random selection and baseline diffusion models by a large margin across different IPC settings. }
    \label{fig:abl_diffusion}
\end{figure}

\subsection{Ablation Study}
% In this section, we conduct extensive ablation study on each component to demonstrate the effectiveness of the proposed minimax fine-tuning scheme. 
% By default the experiments are conducted on ImageWoof. 

\begin{figure*}
    \centering
    \includegraphics[width=0.95\textwidth]{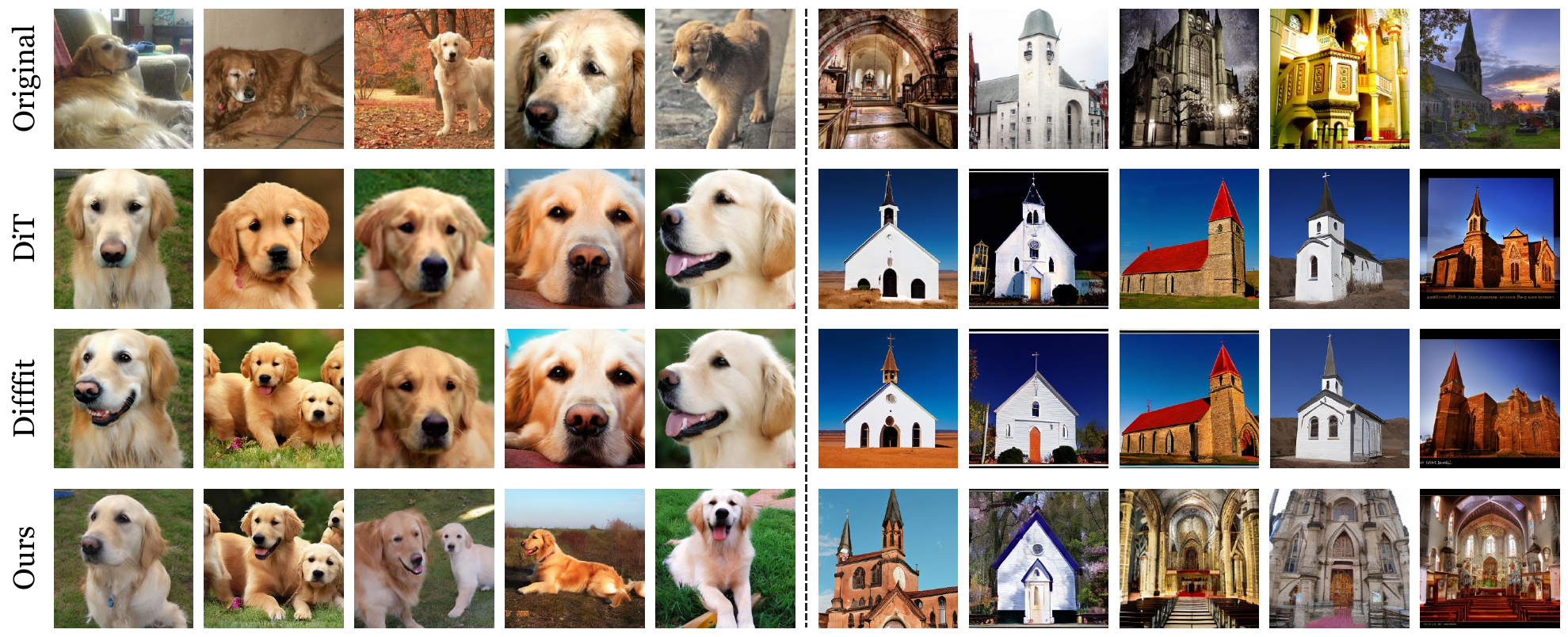}
    \caption{Visualization of random original images, images generated by baseline diffusion models (DiT~\cite{peebles2023scalable} and Difffit~\cite{xie2023difffit}) and our proposed method. For each column, the generated images are based on the same random seed. Comparatively, our method significantly enhances the coverage of original data distribution and the diversity of the surrogate dataset. }
    \label{fig:vis_sample}
\end{figure*}

\noindent
\textbf{Component Analysis.}
We compare the performance with baseline diffusion models to validate the effectiveness of proposed minimax criteria in~\cref{fig:abl_diffusion}. 
The experiments are conducted on ImageWoof and ImageIDC to evaluate the effect on challenging and easy tasks, respectively. 
Under the IPC of 10 and 20, the raw diffusion models (DiT) generate informative images, with validation performance much higher than randomly selected original samples. 
However, as the IPC is continuously increased, the performance gap diminishes for ImageWoof, and random original images surpass the DiT-generated ones at the IPC of 100. 
On ImageIDC the intersection occurs even earlier at the IPC of 50. 
The main reason is reflected in~\cref{fig:tsne-dit}, where the sample diversity remains limited without external guidance. 
The naive Difffit fine-tuning adapts the model to specific domains, yet on large IPCs, the over-fitted generative model still yields inferior performance than the original images. 

\begin{table}[t]
    \centering
    \small
    \caption{The ablation study of the proposed minimax scheme. The result are obtained with ResNet-10 on ImageWoof and ImageIDC. $\mathbf{m}$ denotes the proposed minimax optimization scheme. }
    \label{tab:abl_minimax}
    \setlength{\tabcolsep}{3pt}
    \begin{tabular}{cccc|cccc}
    \toprule
        \multirow{2}{*}{$\mathcal{L}_r$} & $\mathcal{L}_r$ & \multirow{2}{*}{$\mathcal{L}_d$} & $\mathcal{L}_d$ & \multicolumn{2}{c}{ImageWoof} & \multicolumn{2}{c}{ImageIDC} \\
        & w$\backslash\mathbf{m}$ & & w$\backslash\mathbf{m}$ & 10-IPC & 50-IPC & 10-IPC & 50-IPC \\
    \midrule
        - & - & - & - & 35.6$_{\pm 0.9}$ & 51.0$_{\pm 0.9}$ & 53.5$_{\pm 0.2}$ & 66.3$_{\pm 0.2}$ \\
        \checkmark & - & - & - & 34.4$_{\pm 1.1}$  & 47.1$_{\pm 0.5}$ & 49.6$_{\pm 0.7}$ & 60.2$_{\pm 1.2}$ \\
        - & \checkmark & - & - & 37.4$_{\pm 0.4}$  & 49.5$_{\pm 1.0}$ & 54.5$_{\pm 1.2}$ & 65.0$_{\pm 0.8}$ \\
        - & - & \checkmark & - & 35.7$_{\pm 0.8}$  & 48.3$_{\pm 0.6}$ & 51.5$_{\pm 0.6}$ & 64.8$_{\pm 0.8}$ \\
        - & - & - & \checkmark & 38.7$_{\pm 0.9}$  & 54.9$_{\pm 0.7}$ & 52.2$_{\pm 0.6}$ & 68.4$_{\pm 0.7}$ \\
        \checkmark & - & \checkmark & - & 38.3$_{\pm 0.5}$ & 54.9$_{\pm 0.4}$ & 53.3$_{\pm 0.5}$ & 66.8$_{\pm 0.5}$ \\
        - & \checkmark & - & \checkmark & 39.2$_{\pm 1.3}$ & 56.3$_{\pm 1.0}$ & 53.1$_{\pm 0.2}$ & 69.6$_{\pm 0.2}$ \\
    \bottomrule
    \end{tabular}
\end{table}

The addition of representativeness constraint to the training process further enhances the effect of distribution fitting. At small IPCs, the generated images contain richer information, yet for larger IPCs, the lack of diversity brings a negative influence. 
The diversity constraint, in contrast, significantly boosts the information contained in the generated surrogate dataset. 
Despite the performance advantage of $\mathcal{L}_d$ over $\mathcal{L}_r$, combining them still brings stable improvement as our full method. 
Especially on the easier ImageIDC task, grouping these two constraints together contributes to a consistent performance margin over random original images. 
The experimental results validate that both representativeness and diversity play essential parts in constructing effective surrogate datasets. 

\noindent
\textbf{Minimax Scheme.}
In this work, we propose to enhance the representativeness and diversity each with a minimax objective. 
We compare the distillation result with or without the minimax operation in~\cref{tab:abl_minimax}. 
The first row presents the performance of naive Difffit fine-tuning. 
Matching the embeddings to the distribution center as in~\cref{eq:naive_repre} severely degrades the validation performance across all IPCs. 
In contrast, the minimax version constraint as in~\cref{eq:minimax_repre} encourages better coverage, where the performance on small IPCs is improved. 
The effects of diversity constraint and the full method show similar trends. 
The superior performance suggests the effectiveness in enhancing the essential properties of the generative diffusion techniques. 

\begin{figure*}[t]
\begin{subfigure}[t]{0.24\textwidth}
    \centering
    \includegraphics[width=\textwidth]{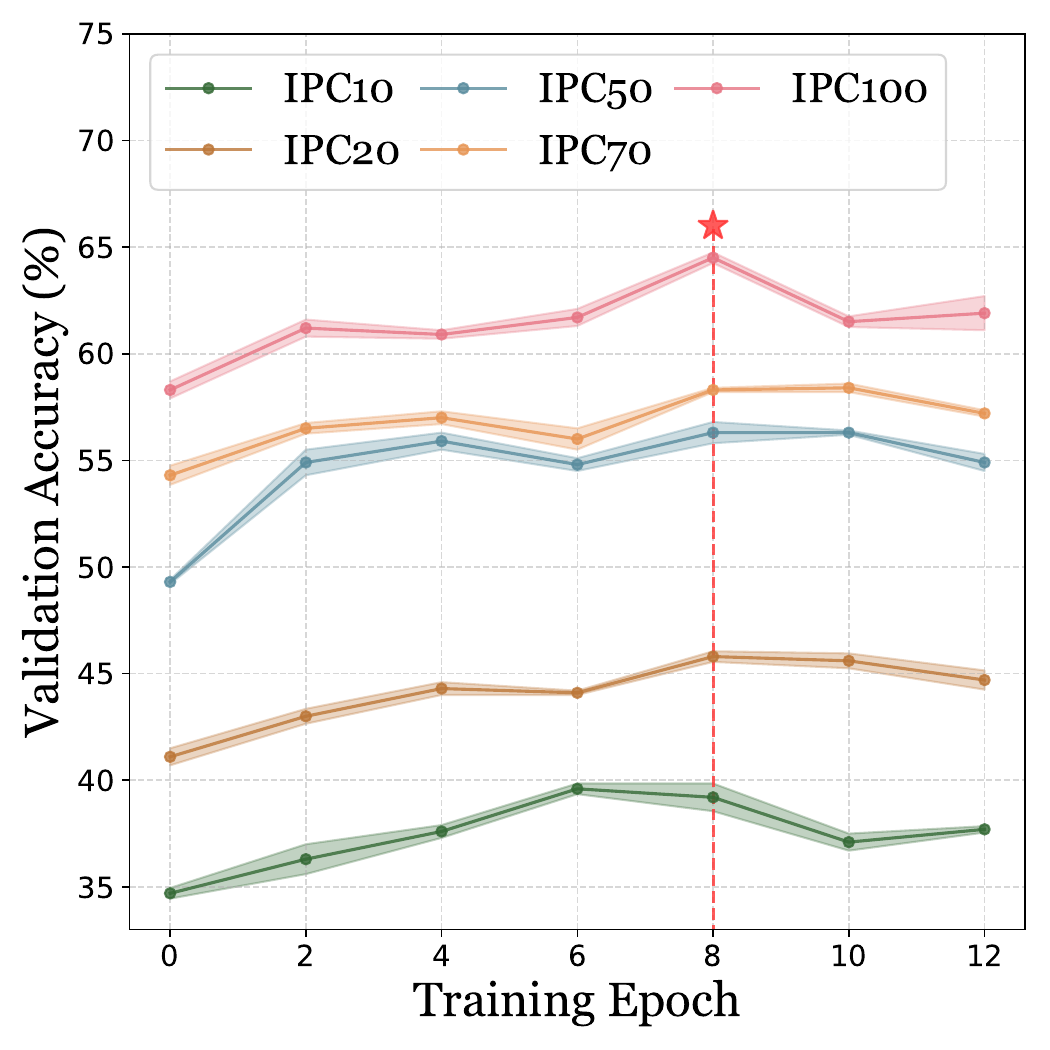}
    \caption{}
    \label{fig:train_curve}
\end{subfigure}
\begin{subfigure}[t]{0.24\textwidth}
    \centering
    \includegraphics[width=\textwidth]{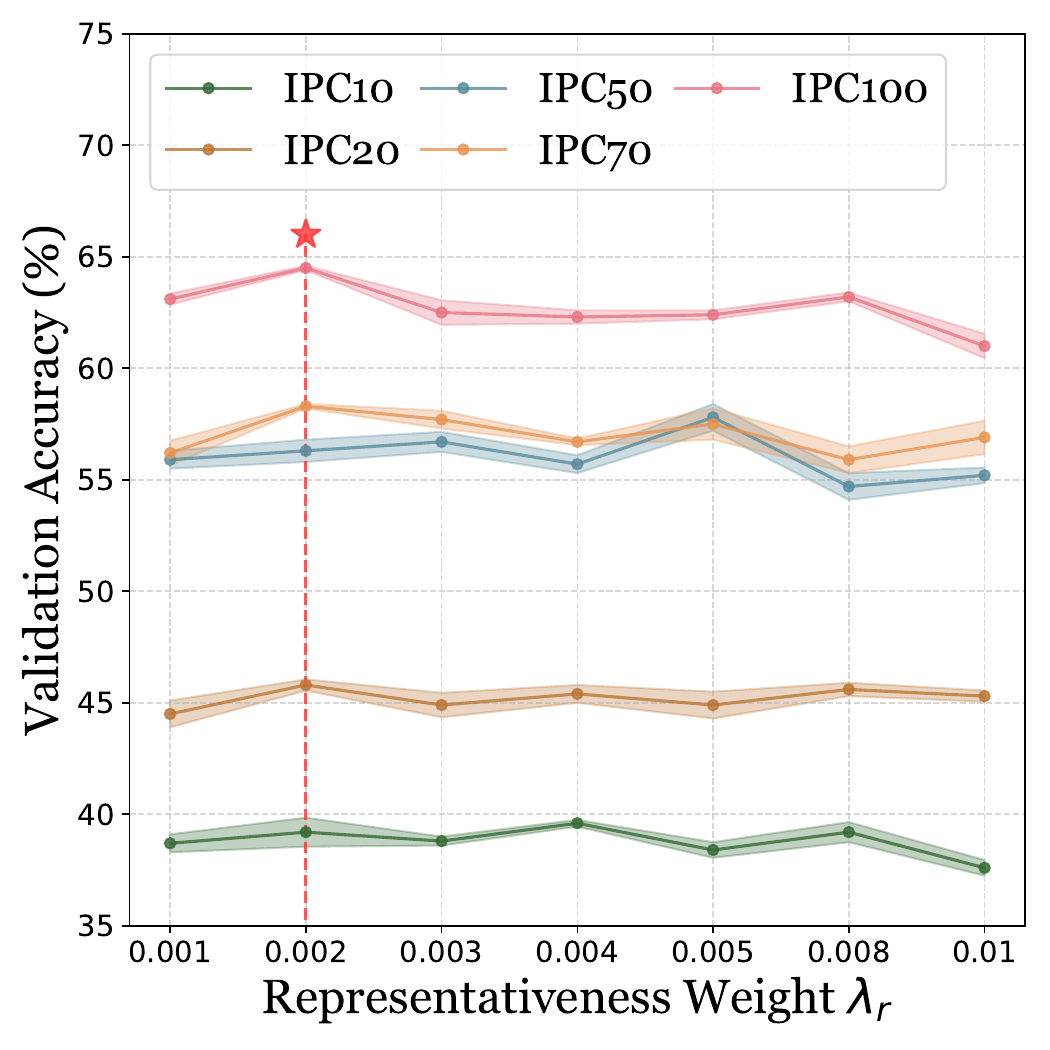}
    \caption{}
    \label{fig:lambda_r}
\end{subfigure}
\begin{subfigure}[t]{0.24\textwidth}
    \centering
    \includegraphics[width=\textwidth]{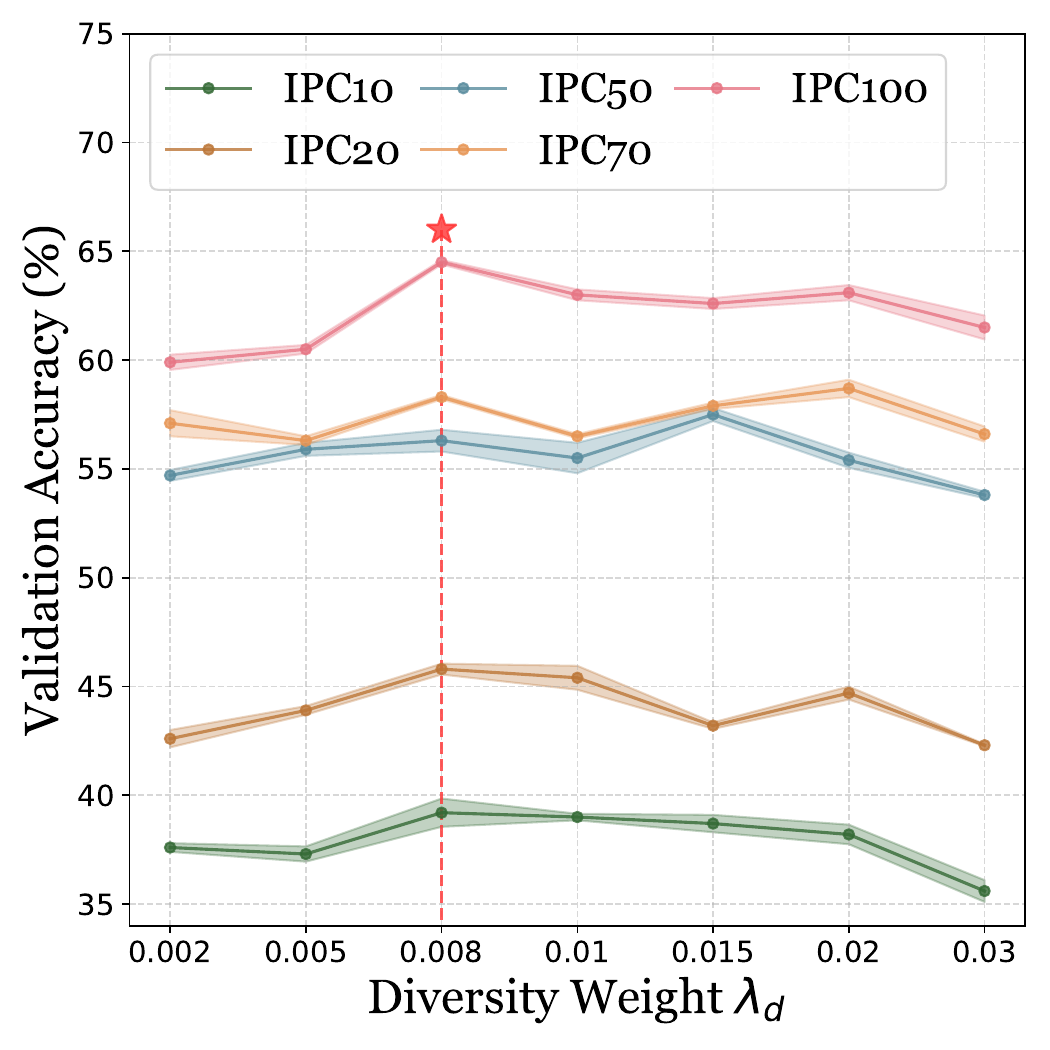}
    \caption{}
    \label{fig:lambda_d}
\end{subfigure}
\begin{subfigure}[t]{0.24\textwidth}
    \centering
    \includegraphics[width=\textwidth]{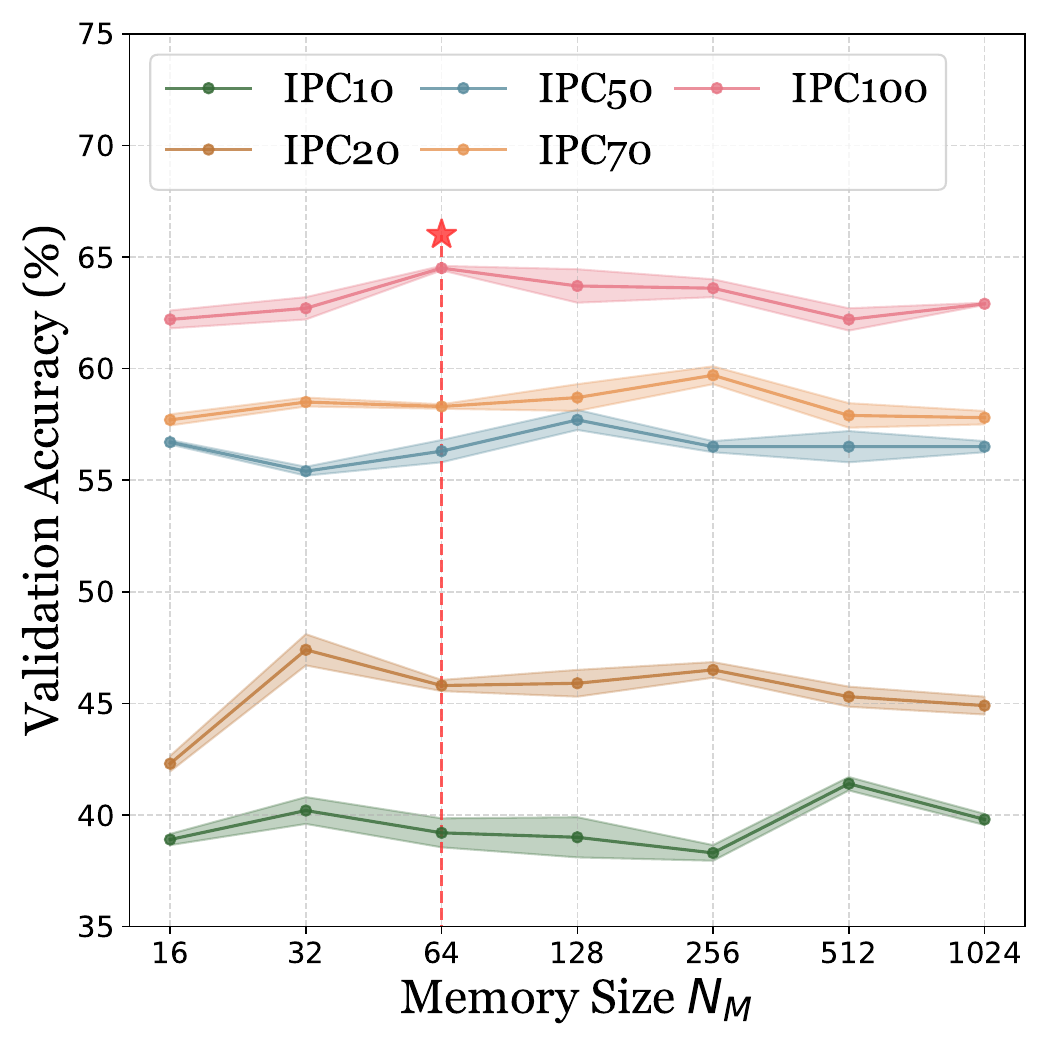}
    \caption{}
    \label{fig:memory}
\end{subfigure}
\caption{Hyper-parameter analysis on (a) the training epochs; (b) the representativeness weight $\lambda_r$; (c) the diversity weight $\lambda_d$; (d) the memory size $N_M$. The results are obtained with ResNetAP-10 on ImageWoof. The dashed line indicates the value adopted in this work. }
\end{figure*}

\subsection{Visualization}

\noindent
\textbf{Sample Distribution Visualization.}
The target of our proposed method is to construct a surrogate dataset with both representativeness and diversity. 
We visualize the t-SNE distribution of samples generated by our proposed method in ~\cref{fig:tsne-dit}. 
In comparison with random original images and baseline diffusion models, our method demonstrates a more thorough coverage over the entire data distribution while maintaining consistency in sample density. 
At the original high-density region, the generated images also form a dense sub-cluster, which is not reflected by random sampling. 
On the other hand, at the original sparse regions, our method exhibits better coverage than baseline diffusion models. 
By simultaneously enhancing the representativeness and diversity in the generative model, the proposed method manages to significantly improve the validation performance of the generated surrogate dataset. 

\noindent
\textbf{Generated Sample Comparison.}
The proposed method notably enhances the representativeness and diversity of the generated surrogate dataset. 
We compare the samples generated with the same random noise (for each column) of different generative methods in~\cref{fig:vis_sample} to explicitly demonstrate the improved properties. 

The images generated by baseline DiT exhibit a realistic high-quality appearance. 
However, the images tend to share similar poses and only present the most prominent features of the objects. 
In the golden retriever case, the generated images mostly present the head part, while for the churches the exterior appearance. 
Difffit fine-tuning further fits the model to the distribution, but in most cases, the differences only lie in small details. 
Comparatively, the proposed minimax criteria significantly enhance both the representativeness and diversity of the generated images. 
On the one hand, there occurs more class-related content in the generated images. 
The golden retriever images include more body parts and the church images encompass the interior layout. 
The minimax optimization leads to better coverage over the entire original distribution, with more related features encapsulated. 
% Through minimax optimization, the generated images better cover the entire original distribution, encapsulating more related features. 
On the other hand, the diversity is significantly enhanced, including variations in pose, background, and appearance styles. 
In such a way the surrogate dataset better represents the original large-scale one, leading to superior validation performance. 
More sample visualizations are provided in the supplementary material. 
% \subsection{Extending to Dataset Expansion}

\noindent
\textbf{Training Curve Visualization.}
We visualize the accuracy curve during the training process in~\cref{fig:train_curve}. 
The validation performance is rapidly improved as the fine-tuning process starts. 
After four epochs, the model tends to converge and reaches the highest performance at the 8th epoch. 
Further extending the training epochs injects excessive diversity into the model, leading to performance degradation. 
We demonstrate the influence of training epochs on the generated images in supplementary material. 

\subsection{Parameter Analysis}
\noindent
\textbf{Objective Weight $\lambda_r$ $\lambda_d$.}
We show the influence of representativeness weight $\lambda_r$ and diversity weight $\lambda_d$ in~\cref{fig:lambda_r} and~\cref{fig:lambda_d}, respectively. 
The $\lambda_r$ variation only produces negligible performance fluctuation on small IPCs, while on large IPCs the performance is also relatively stable. 
For $\lambda_d$, at a proper variation range, the performance is stable. 
However, continuously increasing the diversity of the generated dataset leads to a lack of representativeness, which results in a negative impact.
The negative impact of over-diversity can also validated by the poor performance of K-Center in~\cref{tab:imagenet-10}. 
A uniform performance decrease is observed as $\lambda_d$ reaches 0.03. 
Based on the performance of 100 IPC, we set $\lambda_r$ as 0.002 and $\lambda_d$ as 0.008. 

\noindent
\textbf{Memory Size $N_M$.}
The memory size $N_M$ influences the number of samples involved in the objective calculation. We investigate its influence in~\cref{fig:memory}. 
When the memory is extremely small ($N_M$=16), the provided supervision is also limited, yet the performance is already higher than naive fine-tuning. 
As the memory size is increased in a proper range, the model yields stable performance improvement. 
It is notable that with a larger memory, the performance under the IPC of 10 is better. 
It can be explained by that a larger memory contains more representative information. 
Out of the consideration of performance as well as storage burden, we select the memory of 64 in the other experiments.

\section{Conclusion}

In this work, we propose a novel dataset distillation method based on generative diffusion techniques. 
Through extra minimax criteria, the proposed method significantly enhances the representativeness and diversity of the generated surrogate dataset. 
With much less computational time consumption, the proposed method achieves state-of-the-art validation performance on challenging ImageNet subsets. 
It reduces the resource dependency of previous dataset distillation methods and opens up new possibilities for more practical applications for distilling personalized data. 

\noindent
\textbf{Limitations and Future Works.}
This work mainly focuses on the classification task. 
In future works, we will explore the possibility of incorporating generative techniques for more specific data domains. 

\bibliography{main}
\bibliographystyle{ieeenat_fullname}

\newpage
\clearpage
\setcounter{page}{1}
\maketitlesupplementary

The supplementary material is organized as follows: %Section~\ref{prelim} presents preliminaries; 
Section~\ref{supmat:theory} provides more detailed theoretical analysis; Section~\ref{related-work} presents the related work to this paper; Section~\ref{method pipeline} elaborates upon the method pipeline; Section~\ref{implement} contains additional implementation details; Section~\ref{ablation-supmat} presents some ablation studies; Section~\ref{broader-impact} discusses the broader impact; and finally, Section~\ref{Ethical} presents ethical considerations.

\section{Theoretical Analysis}\label{supmat:theory}

We present the most relevant parts of the referred work~\cite[Section 2.1-2.2]{tzen2019theoretical}. 
Consider that the diffusion takes place over the finite interval $[0,1]$ and let $\mu$ be the desired sample distribution, such that $Z_1\sim \mu$. Assume $\mu$ is absolutely continuous with respect to the standard Gaussian, denoted by $\gamma_d$, and define the Radon-Nikodym derivative $f=d\mu/d\gamma_d$. Then the optimal control, defined in the literature as the F{\" o}llmer drift and expressed as
\[
\begin{array}{l}
u^*(\mathbf{z},t)=\nabla \log Q_{1-t}(f)\\ =\nabla \log\frac{1}{(2\pi(1-t))^{d/2}}\int f(y) \exp\left(-\frac{1}{2(1-t)}\|\mathbf{z}-y\|^2\right)dy
\end{array}
\]
would be such that if $V(Z_t)=u^*(\mathbf{z},t)$ in~\cref{eq:sde}, then this drift would minimize the cost-to-go function:
\[
J^u(\mathbf{z},t):=\mathbb{E}\left[\frac{1}{2}\int_t^1 \|u_s\|^2ds-\log f(Z_1^u)\vert Z_t^u=\mathbf{z}\right].
\]
Equivalently, such a control is the one that, among all such transportation that maps from $\gamma_d$ to $\mu$, minimizes $\int_0^1 \|u_s\|^2ds$ ~\cite{lehec2013representation,eldan2018regularization}. 

The structure of this process presents the opportunity for accurately performing diffusion, enforcing $Z_1^u\sim \mu$, while simultaneously pursuing additional criteria. 
Specifically:
\begin{enumerate}
    \item Immediately we recognize that a nontrivial transportation problem implies the existence of a set (\ie, a nonunique solution to the constraint satisfaction problem) of possible drifts such that the final distribution is $\mu$. We can consider maximizing representativeness as an alternative cost criterion to $\int \|u_s\|^2ds$. To present the criteria in a sensible way, given that
the training is conducted on a minimum across mini-batches, we can instead aim to maximize a bottom quantile, by the cost-to-go functional,
\[
J_r(\mathbf{z},t) = \int_t^1 Q_{\tilde{q},w\sim \mu}\left[\sigma\left(Z_t,w\right)\right]ds,
\]
where $\tilde{q}$ is the quantile percentage, \eg $0.02$ (for instance, if a mini-batch of fifty samples were given, this would be the minimum).
\item Next, notice that with dataset distillation, the small sample size is significant, which suggests that we can consider the aggregate in a particle framework, where for $i=1,...,N_{D}$, we have,
\[
dZ^{u,(i)}_t = u(Z^{u,(i)}_t,t) dt+dW_t,\, t\in [0,1];\, Z^{u,(i)}_0=\mathbf{z}_0
\]
%with collective distribution,
%\[
%\rho^{\tilde D}(t) = \frac{1}{N_D}\sum\limits_{i=1}^{N_D} \delta_{X^{u,(i)}_t}
%\]
presenting an additional degree of freedom, which we take advantage of by encouraging diversity, \ie, minimizing
\[
J_d(\mathbf{z},1) =  \max\limits_{i,j=1,..,N_D} \sigma\left(Z_1^{u,(i)},Z_1^{u,(j)}\right).
\]
 Since generation accuracy and representativeness are criteria for individual particles, maximizing diversity across particles can be considered as optimizing with respect to the additional degree of freedom introduced by having multiple particles.
\end{enumerate}
Thus, we can see that it presents the opportunity to consider generative diffusion as a bi-level stochastic control problem. 

A brief note on convergence guarantees for~\cref{eq:finitecontrol} presented in the main paper. A straightforward extension of~\cite{doan2022nonlinear} to three layers (similar to the extension from bi-level to tri-level convex optimization in~\cite{shafiei2021trilevel}) yields convergence guarantees in expectation to a stationary point for all objectives. 
It is important to note that in the case of nonconvex objectives, the asymptotic point will satisfy a fairly weak condition. Specifically, it may not be stationary for the top objective, as the lower levels are not necessarily at global minimizers. This is, however, the best that can be ensured with stochastic gradient based methods and similar.

\section{Related Works} \label{related-work}

\subsection{Dataset Distillation}
Dataset distillation (DD) aims to condense the information of large-scale datasets into small amounts of synthetic images with close training performance~\cite{wang2018dataset,zhao2020dataset,kim2022dataset,cui2022dc}. 
The informative images are also useful for tasks like continual learning~\cite{kim2022dataset,gu2023summarizing}, federated learning~\cite{liu2022MetaKnowledgeCondensation,xiong2023FedDMIterativeDistribution} and neural architecture search~\cite{such2020GenerativeTeachingNetworks}. 
Previous DD works can be roughly divided into bi-level optimization and training metric matching methods. 
Bi-level optimization methods incorporate meta learning into the surrogate image update~\cite{nguyen2021dataset,nguyen2021datasetkrr,zhou2022dataset,loo2022efficient,loo2023dataset,deng2022RememberDistillingDatasets}. 
In comparison, metric matching methods optimize the synthetic images by matching the training gradients~\cite{zhao2020dataset,zhao2021dataset,kim2022dataset,liu2023dream,lee2022dataset,vahidian2024group}, feature distribution~\cite{zhao2023dataset,wang2022cafe,sajediDataDAMEfficientDataset,zhao2023ImprovedDistributionMatching}, predicted logits~\cite{wang2023dim} or training trajectories~\cite{wu2023multimodal,cazenavette2022dataset,du2023MinimizingAccumulatedTrajectory} with original images. 

\subsection{Data Generation with Diffusion}
The significantly improved image quality and sample diversity by diffusion models opens up new possibilities for data generation~\cite{ho2020DenoisingDiffusionProbabilistic,dhariwalDiffusionModelsBeat2021,kingmaVariationalDiffusionModels2021,nicholImprovedDenoisingDiffusion2021}. 
Through prompt engineering~\cite{sariyildiz2023FakeItTill,dunlap2023diversify,he2022SYNTHETICDATAGENERATIVE}, latent interpolation~\cite{zhou2023training} and classifier-free guidance~\cite{zhou2023training,azizi2023synthetic}, the diversity-improved synthetic images are useful to serve as augmentation or expansion for the original samples. 
The generated images also contribute to zero-shot image classification tasks~\cite{shipard2023DiversityDefinitelyNeeded}. 
However, these works mainly focus on recovering the original distribution with equal or much larger amounts of data.
In contrast, we intend to distill the rich data information into small surrogate datasets. 
Moreover, prompt engineering usually requires special designs according to different data classes, while our proposed method saves extra effort. 
As far as we have investigated, there are no previous attempts to incorporate generative diffusion techniques into the dataset distillation task. 
In addition to diffusion models, there are also some previous works considering the diversity issue for Generative Adversarial Networks (GANs)~\cite{besnier2020ThisDatasetDoes,gurumurthy2017DeLiGANGenerativeAdversarial,mao2019ModeSeekingGenerative,yang2018DiversitySensitiveConditionalGenerative}. 
However, the improvement in diversity is not reflected in downstream tasks. 
In this work, we seek to enhance both representativeness and diversity for constructing a small surrogate dataset with similar training performance compared with original large-scale ones. 

\begin{figure}[t]
    \centering
    \includegraphics[width=\columnwidth]{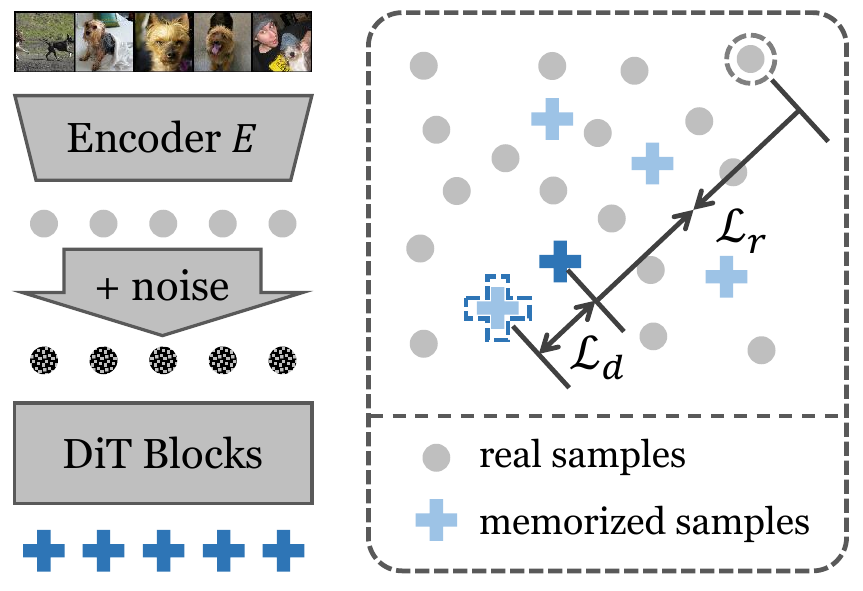}
    \caption{The training pipeline of the proposed minimax diffusion fine-tuning. The DiT blocks predict the added noise and original embeddings (dark-blue crossings). Then the parameters are updated with the simple diffusion objective and the minimax objectives. The minimax objectives (the right part) enforce the predicted embedding to be close to the farthest real sample and be far away from the closest predicted embedding of adjacent iterations. }
    \label{fig:framework}
\end{figure}

\section{Method Pipeline} \label{method pipeline}
We demonstrate the pipeline of the proposed minimax fine-tuning method in Fig.~\ref{fig:framework}. 
The real images are first passed through the encoder $E$ to obtain the original embeddings $\mathbf{z}$. 
Random noise $\epsilon$ is then added to the embeddings by the diffusion process. 
The DiT blocks then predict the added noise, with which the predicted original embeddings $\hat{\mathbf{z}}$ (dark-blue crossings in~\cref{fig:framework}) are also able to be calculated. 
We maintain two auxiliary memories $\mathcal{M}$ (grey dots) and $\mathcal{D}$ (light-blue crossings) to store the encountered real embeddings and predicted embeddings at adjacent iterations, respectively. 
The denoised embeddings of the current iteration are pushed away from the most similar predicted embedding and are pulled close to the least similar real embedding. 
The DiT blocks are optimized with the proposed minimax criteria and the simple diffusion training loss $\mathcal{L}_{simple}$ as in~\cref{eq:simple}. 

At the inference stage, given a random noise together with a specified class label, the DiT network predicts the noise that requires to be subtracted. 
Then the Decoder $D$ recovers the images from the denoised embeddings.

\section{More Implementation Details} \label{implement}
We conduct experiments on three commonly adopted network architectures in the area of DD, including:
\begin{enumerate}
    \item \textbf{ConvNet-6} is a 6-layer convolutional network. In previous DD works where small-resolution images are distilled, the most popular network is ConvNet-3~\cite{kim2022dataset,zhao2021dataset,liu2023dream}. 
    We extend an extra 3 layers for full-sized 256$\times$256 ImageNet data. 
    The network contains 128 feature channels in each layer, and instance normalization is adopted. 
    \item \textbf{ResNetAP-10} is a 10-layer ResNet~\cite{he2016deep}, where the strided convolution is replaced by average pooling for downsampling. 
    \item \textbf{ResNet-18} is a 18-layer ResNet~\cite{he2016deep} with instance normalization (IN). As the IN version performs better than batch normalization under our protocol, we uniformly adopt IN for the experiments. 
\end{enumerate}

For diffusion fine-tuning, an Adam optimizer is adopted with the learning rate set as 0.001, which is consistent with the original Difffit setting~\cite{xie2023difffit}. 
We set the mini-batch size as 8 mainly due to the GPU memory limitation. 
The employed augmentations during the fine-tuning stage include random resize-crop and random flip. 

\begin{table}[t]
    \centering
    \caption{The training epoch number on different IPC settings for distilled dataset validation. }
    \label{tab:epoch_valid}
    \begin{tabular}{l|ccccc}
    \toprule
        IPC & 10 & 20 & 50 & 70 & 100 \\
        \midrule
        Epochs & 2000 & 1500 & 1500 & 1000 & 1000 \\
        \bottomrule
    \end{tabular}
\end{table}

\begin{table*}[t]
    \centering
    \caption{Performance comparison on ImageNet-100. The best results are marked as \textbf{bold}. }
    \label{tab:imagenet-100}
    \begin{tabular}{ll|ccc|c|c}
    \toprule
        IPC (Ratio) & Test Model & Random & Herding~\cite{welling2009herding} & IDC-1~\cite{kim2022dataset} & Ours & Full \\
    \midrule
        \multirow{3}{*}{10 (0.8\%)}
        & ConvNet-6 & 17.0$_{\pm 0.3}$ & 17.2$_{\pm 0.3}$ & \textbf{24.3$_{\pm 0.5}$} & 22.3$_{\pm 0.5}$ & 79.9$_{\pm 0.4}$ \\
        & ResNetAP-10 & 19.1$_{\pm 0.4}$ & 19.8$_{\pm 0.3}$ & \textbf{25.7$_{\pm 0.1}$} & 24.8$_{\pm 0.2}$ & 80.3$_{\pm 0.2}$ \\
        & ResNet-18 & 17.5$_{\pm 0.5}$ & 16.1$_{\pm 0.2}$ & \textbf{25.1$_{\pm 0.2}$} & 22.5$_{\pm 0.3}$ & 81.8$_{\pm 0.7}$ \\
    \midrule
        \multirow{3}{*}{20 (1.6\%)}
        & ConvNet-6 & 24.8$_{\pm 0.2}$ & 24.3$_{\pm 0.4}$ & 28.8$_{\pm 0.3}$ & \textbf{29.3$_{\pm 0.4}$} & 79.9$_{\pm 0.4}$ \\
        & ResNetAP-10 & 26.7$_{\pm 0.5}$ & 27.6$_{\pm 0.1}$ & 29.9$_{\pm 0.2}$ & \textbf{32.3$_{\pm 0.1}$} & 80.3$_{\pm 0.2}$ \\
        & ResNet-18 & 25.5$_{\pm 0.3}$ & 24.7$_{\pm 0.1}$ & 30.2$_{\pm 0.2}$ & \textbf{31.2$_{\pm 0.1}$} & 81.8$_{\pm 0.7}$ \\
    \bottomrule
    \end{tabular}
\end{table*}

\begin{table}[t]
    \centering
    \caption{The influence of diffusion denoising step number on the generation time of each image and the corresponding validation performance. Performance evaluated with ResNet-10 on ImageWoof. The best results are marked as \textbf{bold}. }
    \label{tab:denoise_step}
    \begin{tabular}{lc|ccc}
        \toprule
        &  & \multicolumn{3}{c}{Denoising Step} \\
        &  & 50 & 100 & 250 \\
        \midrule
        \multicolumn{2}{c|}{Time (s)} & 0.8 & 1.6 & 3.2 \\
        \midrule
        \parbox[t]{2mm}{\multirow{5}{*}{\rotatebox[origin=c]{90}{IPC}}}
        & 10 & 39.2$_{\pm 1.3}$ & 35.7$_{\pm 0.7}$ & \textbf{39.6$_{\pm 0.9}$} \\
        & 20 & \textbf{45.8$_{\pm 0.5}$} & 44.5$_{\pm 0.6}$ & 43.7$_{\pm 0.7}$ \\
        & 50 & 56.3$_{\pm 1.0}$ & \textbf{58.4$_{\pm 0.5}$} & 55.8$_{\pm 0.5}$ \\
        & 70 & 58.3$_{\pm 0.2}$ & \textbf{59.6$_{\pm 1.1}$} & 58.9$_{\pm 1.4}$ \\
        & 100 & \textbf{64.5$_{\pm 0.2}$} & 63.3$_{\pm 0.7}$ & 62.8$_{\pm 0.6}$ \\
        \bottomrule
    \end{tabular}
\end{table}

\begin{figure*}
    \centering
    \includegraphics[width=\textwidth]{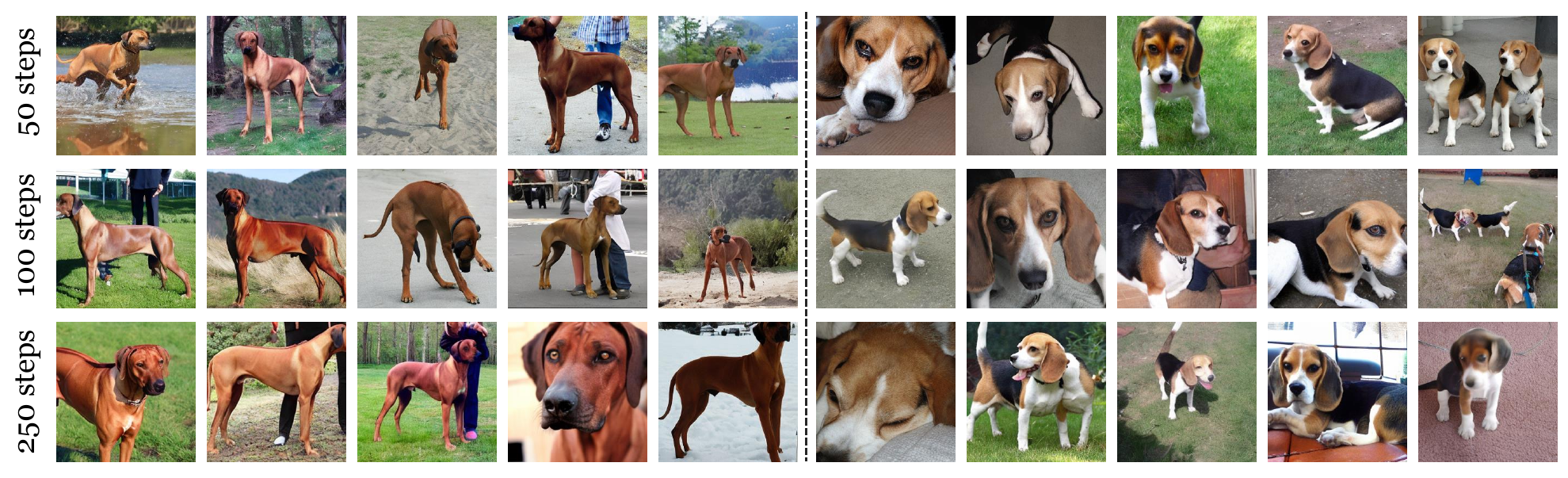}
    \caption{Visualization of images generated by the same model with different denoising steps. For each column, the generated images are based on the same random seed.  }
    \label{fig:vis_sample_step}
\end{figure*}

For the validation training, we adopt the same protocol as in~\cite{kim2022dataset}. 
Specifically, a learning rate of 0.01 for an Adam optimizer is adopted. 
The training epoch setting is presented in~\cref{tab:epoch_valid}. 
The reduced training epochs also partly explain the reason why the performance gap between the IPC settings of 50 and 70 is relatively small. 
The adopted data augmentations include random resize-crop and CutMix. 

\begin{figure*}[t]
\begin{subfigure}[t]{0.24\textwidth}
    \centering
    \includegraphics[width=\textwidth]{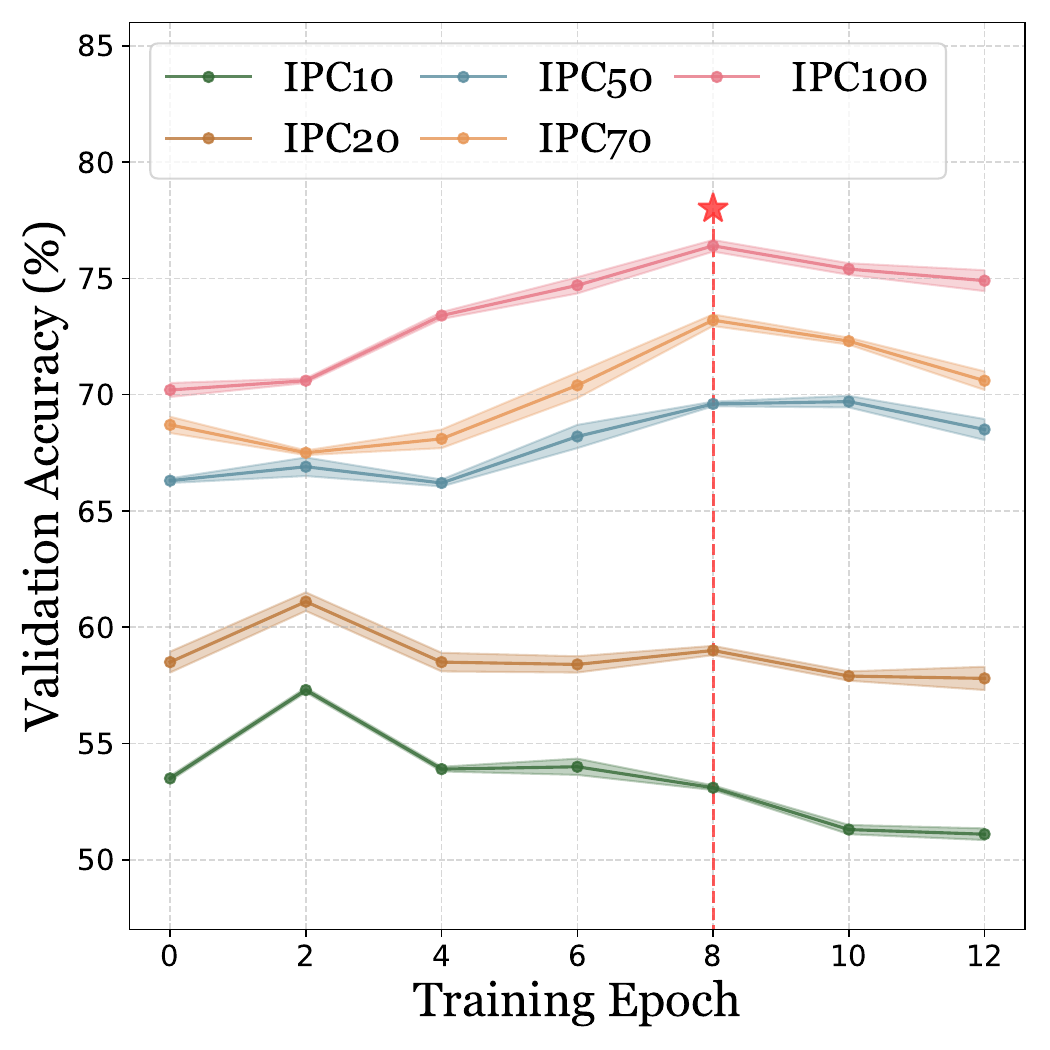}
    \caption{}
    \label{fig:train_curve_idc}
\end{subfigure}
\begin{subfigure}[t]{0.24\textwidth}
    \centering
    \includegraphics[width=\textwidth]{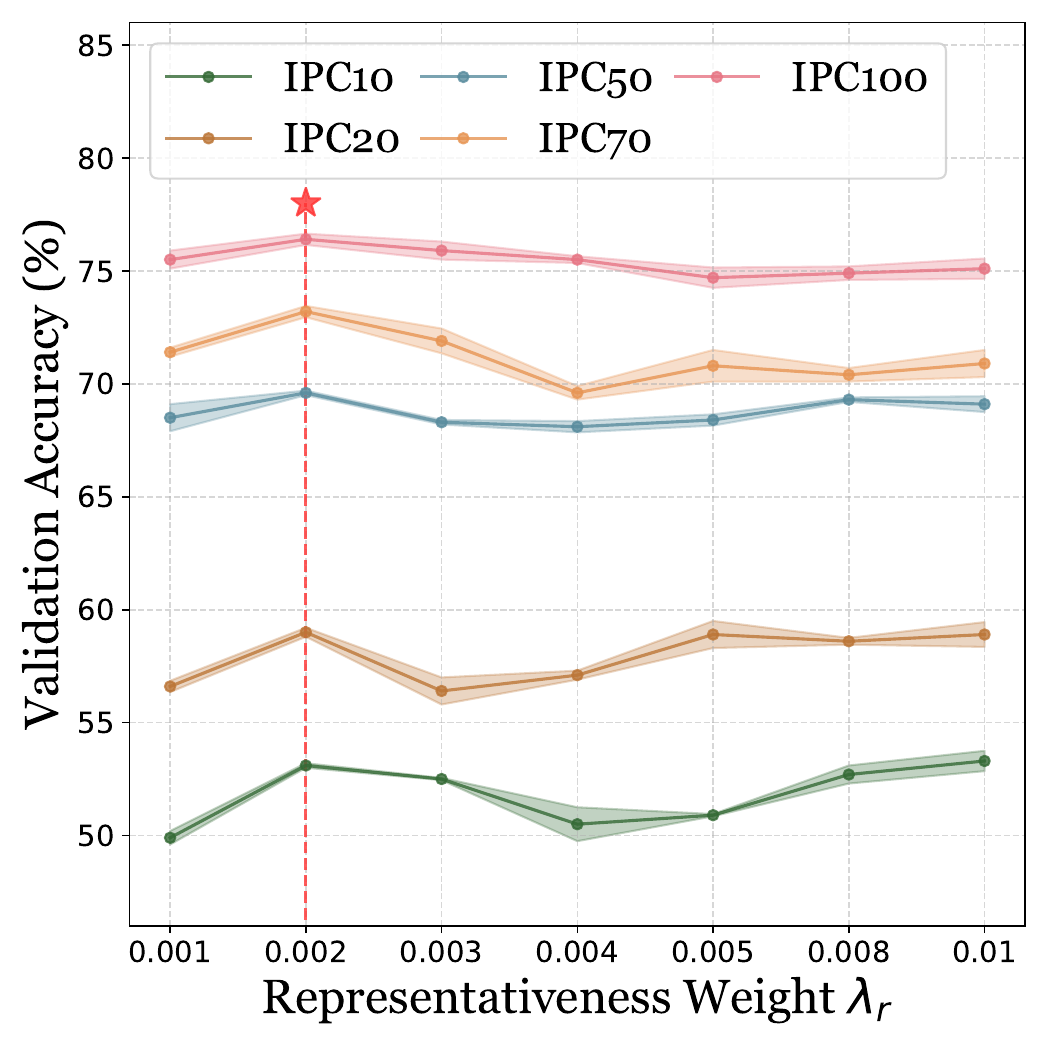}
    \caption{}
    \label{fig:lambda_r_idc}
\end{subfigure}
\begin{subfigure}[t]{0.24\textwidth}
    \centering
    \includegraphics[width=\textwidth]{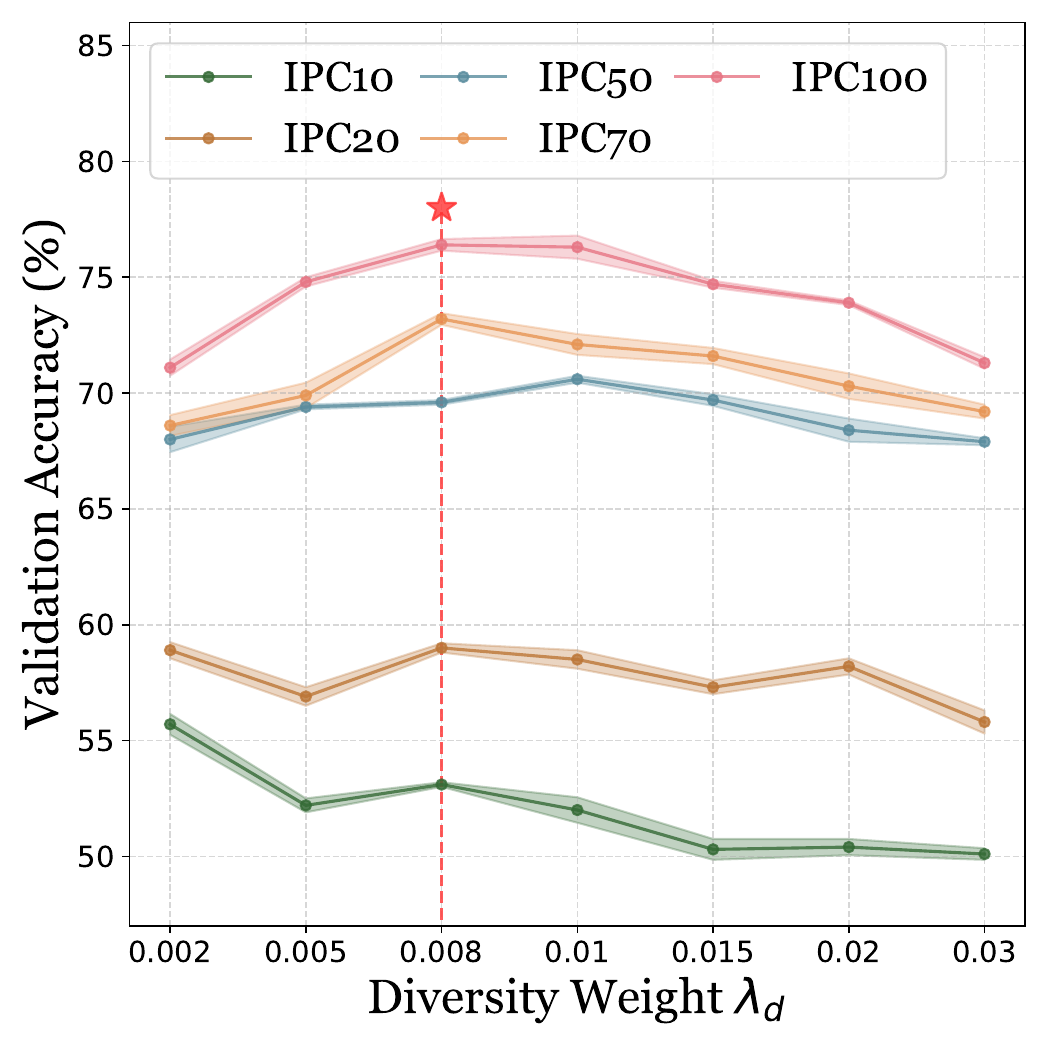}
    \caption{}
    \label{fig:lambda_d_idc}
\end{subfigure}
\begin{subfigure}[t]{0.24\textwidth}
    \centering
    \includegraphics[width=\textwidth]{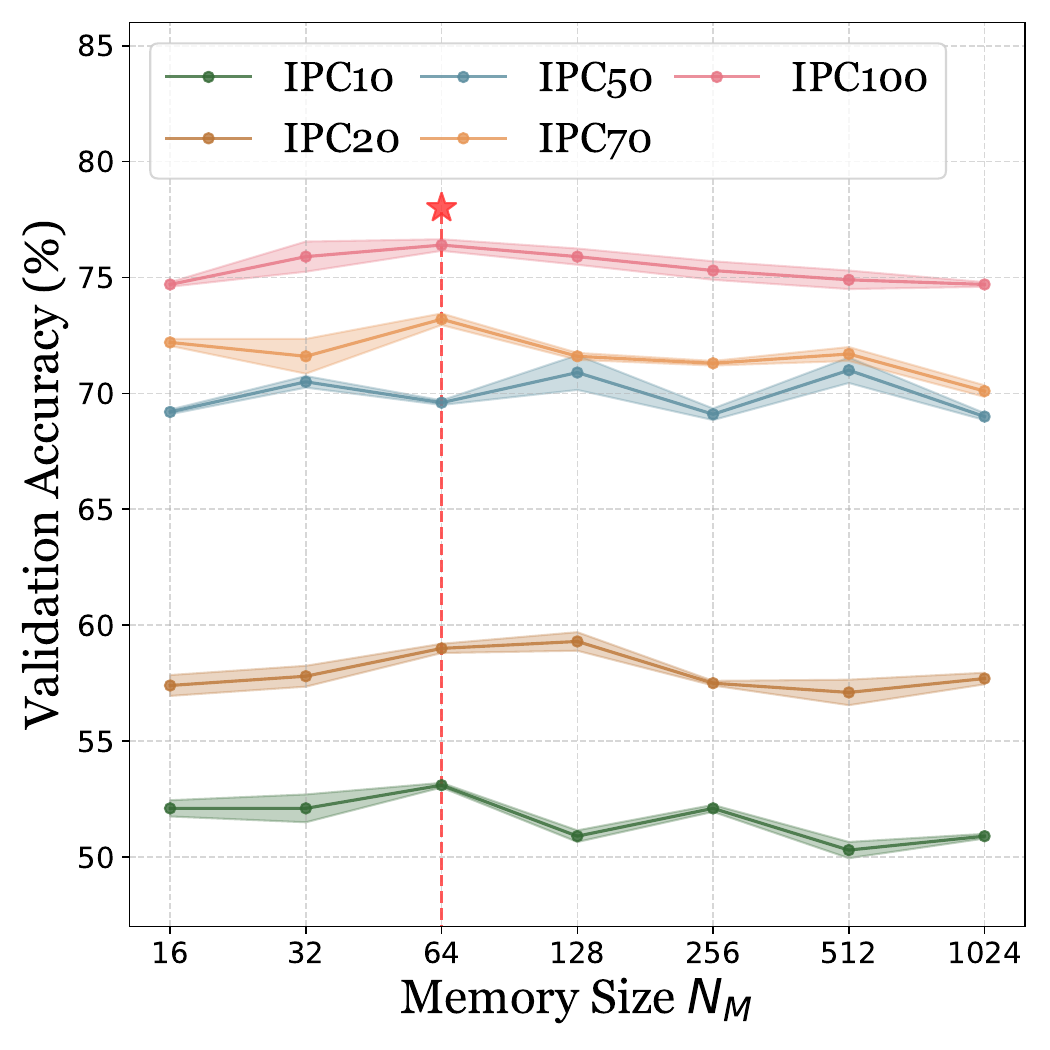}
    \caption{}
    \label{fig:memory_idc}
\end{subfigure}
\caption{Hyper-parameter analysis on (a) the training epochs; (b) the representativeness weight $\lambda_r$; (c) the diversity weight $\lambda_d$; (d) the memory size $N_M$. The results are obtained with ResNetAP-10 on ImageIDC. The dashed line indicates the value adopted in this work. }
\end{figure*}

\begin{figure*}
    \centering
    \includegraphics[width=\textwidth]{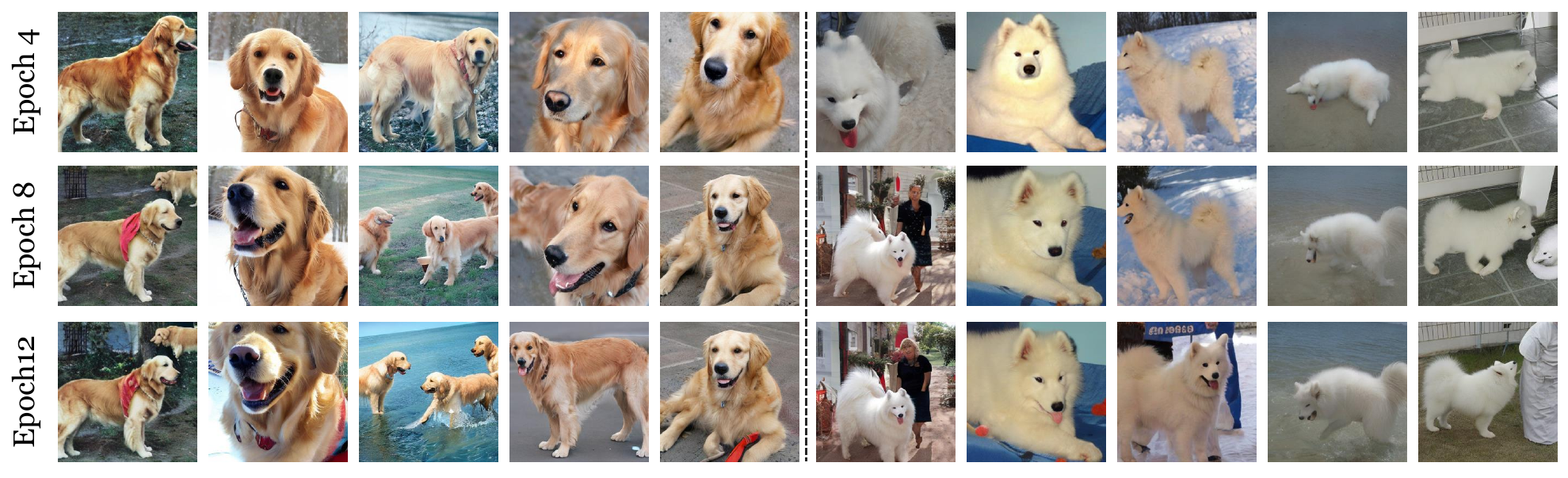}
    \caption{Visualization of images generated by models after different epochs of training. For each column, the images are generated based on the same random noise. }
    \label{fig:vis_sample_epoch}
\end{figure*}

\section{More Analysis and Discussion} \label{ablation-supmat}
\subsection{Experiments to ImageNet-100}
In addition to the 10-class ImageNet subsets and full ImageNet-1K, we also conduct experiments on ImageNet-100, and the results are shown in~\cref{tab:imagenet-100}. 
The validation protocol follows that in IDC~\cite{kim2022dataset}.
Due to the limitation of computational resources, here we directly employ the official distilled images of IDC-1~\cite{kim2022dataset} for evaluation. 
The original resolution is 224$\times$224, and we resize the images to 256$\times$256 for fair comparison. 
Under the IPC setting of 10, IDC-1 achieves the best performance. 
Yet when the IPC increases, the performance gap between the distilled images of IDC-1 and randomly selected original images is smaller. 
Comparatively, our proposed minimax diffusion method consistently provides a stable performance improvement over original images across different IPC settings. 
\emph{It is worth noting that for IDC-1, the distillation process on ImageNet-100 demands hundreds of hours, while the proposed minimax diffusion only requires \textbf{10 hours}}. 
The significantly reduced training time offers much more application possibilities for the dataset distillation techniques. 

\subsection{Diffusion Denoising Step}
In our experiments, we set the diffusion denoising step number as 50. 
We evaluate its influence on the validation performance in Tab.~\ref{tab:denoise_step}. 
There are no fixed patterns for achieving better performance across all the IPCs. 
Additionally, we compare the generated images under different step settings in Fig.~\ref{fig:vis_sample_step}. 
For DiT~\cite{peebles2023scalable}, the denoising process is conducted in the embedding space. 
Therefore, it is reasonable that with different steps the generated images are variant in the pixel space. 
It can be observed that under all steps, the model generates high-quality images with sufficient diversity. 
Taking the calculation time into consideration, we simply select 50 steps in our experiments. 

\subsection{Parameter Analysis on ImageIDC}
We extensively demonstrate the parameter analysis on ImageIDC to illustrate the robustness of the hyper-parameters. 
\cref{fig:train_curve_idc} shows the performance curve along the training epochs. 
As the training process starts, the representativeness constraint quickly improves the accuracy of small IPCs. 
Further training enhances the diversity, where the performance on large and small IPCs shows different trends. 
Generally, the generated images achieve the best performance at the 8th epoch, which is consistent with the ImageWoof experiments. 

\begin{table}[t]
    \centering
    \caption{The dataset expansion results of the 100-IPC generated images on ImageWoof.}
    \label{tab:expansion}
    \begin{tabular}{l|cc}
    \toprule
        Test Model & Original & Original + 100-IPC \\
    \midrule
        ConvNet-6 & 86.4$_{\pm 0.2}$ & \textbf{87.0$_{\pm 0.6}$} \\
        ResNetAP-10 & 87.5$_{\pm 0.5}$ & \textbf{89.3$_{\pm 0.6}$} \\
        ResNet-18 & 89.3$_{\pm 1.2}$ & \textbf{90.1$_{\pm 0.3}$} \\
    \bottomrule
    \end{tabular}
\end{table}

Compared with the results on ImageWoof, further enlarging the representativeness weight $\lambda_r$ improves the performance on small IPCs, as illustrated in~\cref{fig:lambda_r_idc}. 
In comparison, increasing diversity causes a drastic performance drop in~\cref{fig:lambda_r_idc}. 
Although the default settings remain relatively better choices, the balance point between representativeness and diversity is worthy of further exploration. 
The memory size $N_M$ merely has a mild influence on the performance, which aligns with that of ImageWoof. 

\subsection{Extension to Dataset Expansion}
In addition to the standard dataset distillation task, where a small surrogate dataset is generated to replace the original one, we also evaluate the capability of the generated images as an expanded dataset. 
We add the generated 100-IPC surrogate dataset to the original ImageWoof (approximately 1,300 images per class) and conduct the validation in~\cref{tab:expansion}. 
As can be observed, although the extra images only take up a small ratio compared with the original data, a considerable performance improvement is still achieved. 
\emph{The results support that the proposed minimax diffusion can also be explored as a dataset expansion method in future works. }

\subsection{Generated Samples of Different Epochs}
We visualize the images generated by models after different epochs of training in Fig.~\ref{fig:vis_sample_epoch} to explicitly demonstrate the training effect of the proposed minimax diffusion method. 
As the training proceeds, the generated images present variation trends from several perspectives. 
Firstly, the images tend to have more complicated backgrounds and environments, such as more realistic water and objects of other categories (\textit{e.g.} human). 
Secondly, there are more details filled in the images, like the clothes in the first column and the red spots in the sixth. 
These new facets significantly enhance the diversity of the generated surrogate dataset. 
Furthermore, through the fine-tuning process, the class-related features are also enhanced. 
In the ninth and tenth columns, the model at the fourth epoch fails to generate objects with discriminative features. 
In comparison, the images generated by subsequent models demonstrate substantial improvement regarding the representativeness property. 

\begin{table}[t]
    \centering
    \caption{The averaged generation quality evaluation of 10 classes each with 100 images in ImageWoof. }
    \label{tab:fid}
    \small
    \setlength{\tabcolsep}{4pt}
    \begin{tabular}{c|cccc}
    \toprule
        Method & FID & Precision (\%) & Recall (\%) & Coverage (\%) \\
    \midrule
        DM~\cite{zhao2023dataset} & 208.6 & 22.1 & 23.8 & 5.8 \\
        DiT~\cite{peebles2023scalable} & 81.4 & 92.8 & 38.9 & 24.1 \\
        DiT+$\mathcal{L}_r$ & 85.4 & 93.2 & 38.1 & 24.6 \\
        DiT+$\mathcal{L}_d$ & 81.1 & 90.4 & 46.8 & 28.3 \\
        Ours full & 81.5 & 92.4 & 45.3 & 28.6 \\
    \bottomrule
    \end{tabular}
\end{table}

\noindent
\subsection{Generation Quality Evaluation. }
We further report quantitative evaluations on the generation quality by adding the proposed minimax criteria in~\cref{tab:fid}. 
The representativeness and diversity constraints improve the precision and recall of the generated data, respectively. 
The full method finds a balanced point between these two properties while obtaining the best coverage over the whole distribution. 
The fine-tuning brings negligible influence on the FID metric. 
And all the metrics of our proposed method are significantly better than those attained by DM~\cite{zhao2023dataset}. 

\subsection{Generated Samples of Different Classes}
\begin{figure*}
    \centering
    \includegraphics[width=0.98\textwidth]{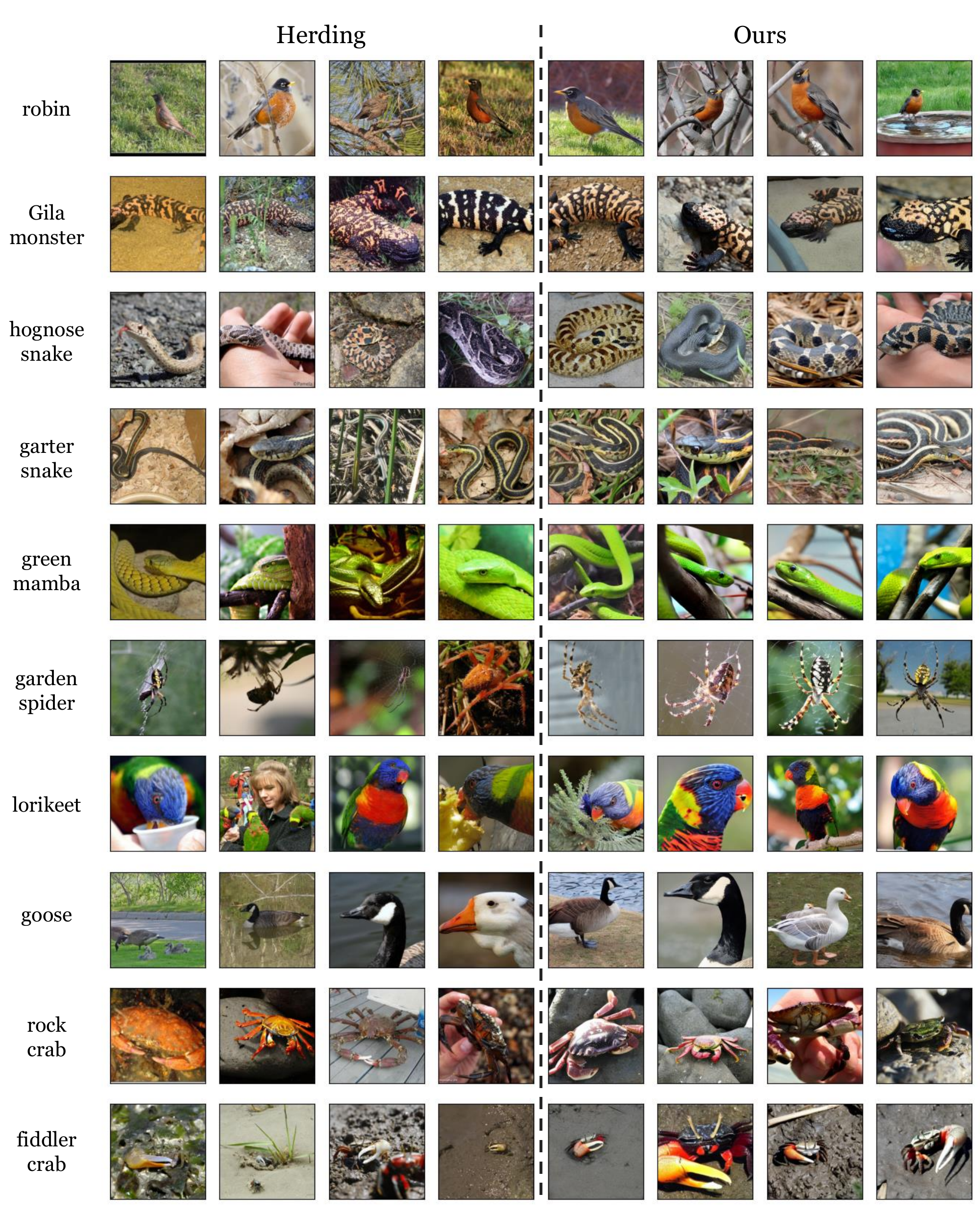}
    \caption{Comparison between samples selected by Herding (left) and generated by the proposed minimax diffusion method (right) for ImageNet-100 classes 0-9. The class names are marked at the left of each row. }
    \label{fig:phase0}
\end{figure*}

\begin{figure*}
    \centering
    \includegraphics[width=0.98\textwidth]{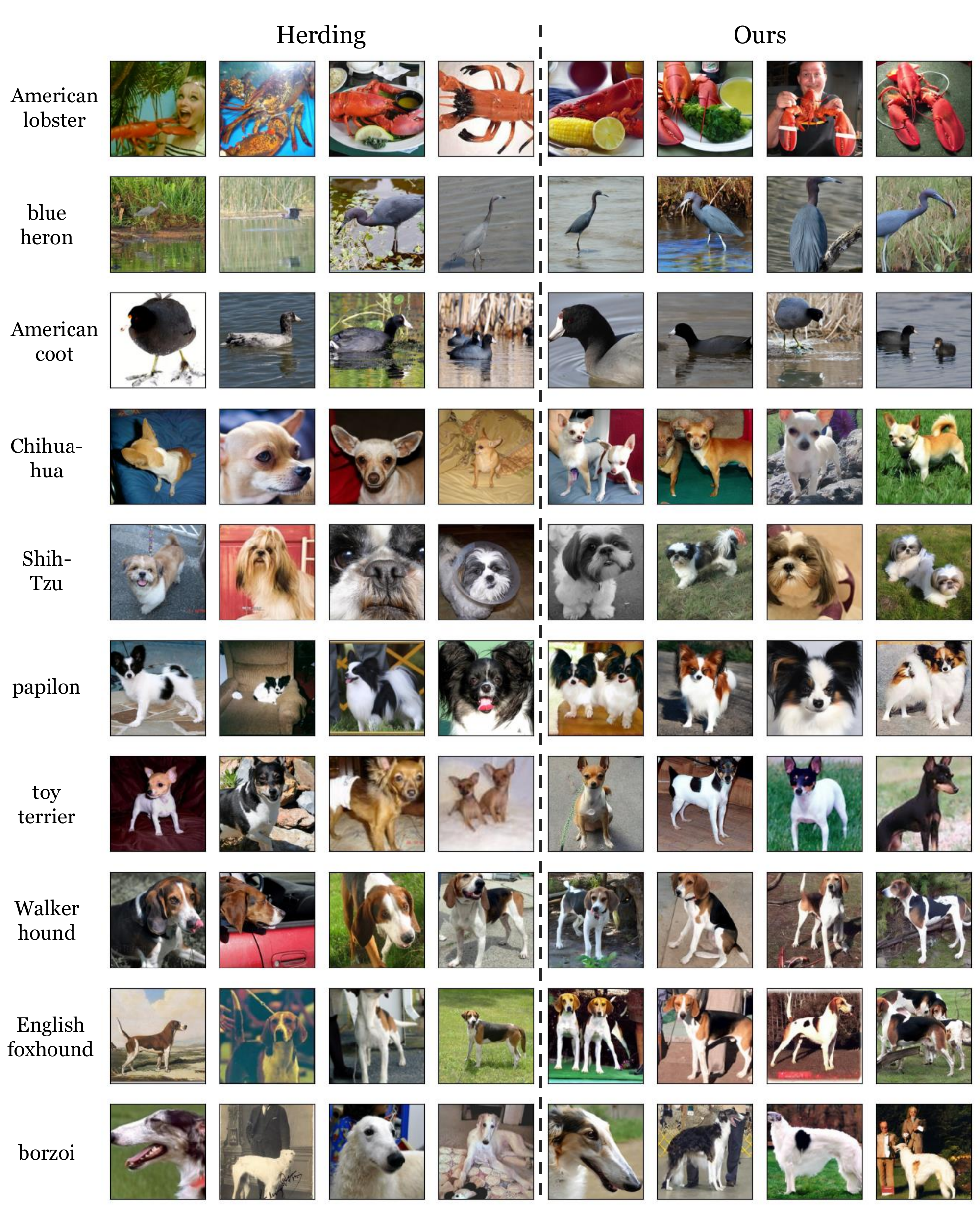}
    \caption{Comparison between samples selected by Herding (left) and generated by the proposed minimax diffusion method (right) for ImageNet-100 classes 10-19. The class names are marked at the left of each row. }
    \label{fig:phase1}
\end{figure*}

\begin{figure*}
    \centering
    \includegraphics[width=0.98\textwidth]{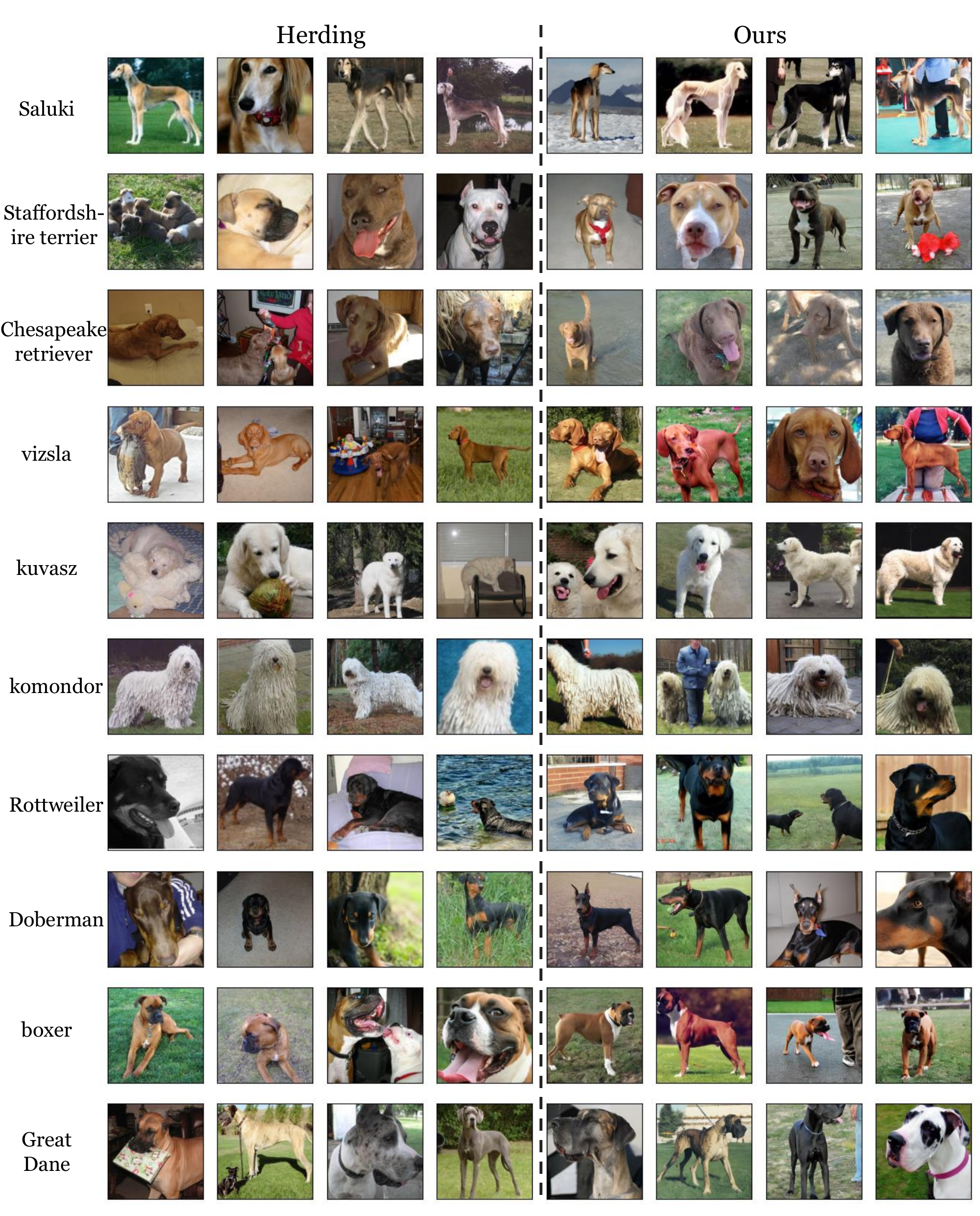}
    \caption{Comparison between samples selected by Herding (left) and generated by the proposed minimax diffusion method (right) for ImageNet-100 classes 20-29. The class names are marked at the left of each row. }
    \label{fig:phase2}
\end{figure*}

\begin{figure*}
    \centering
    \includegraphics[width=0.98\textwidth]{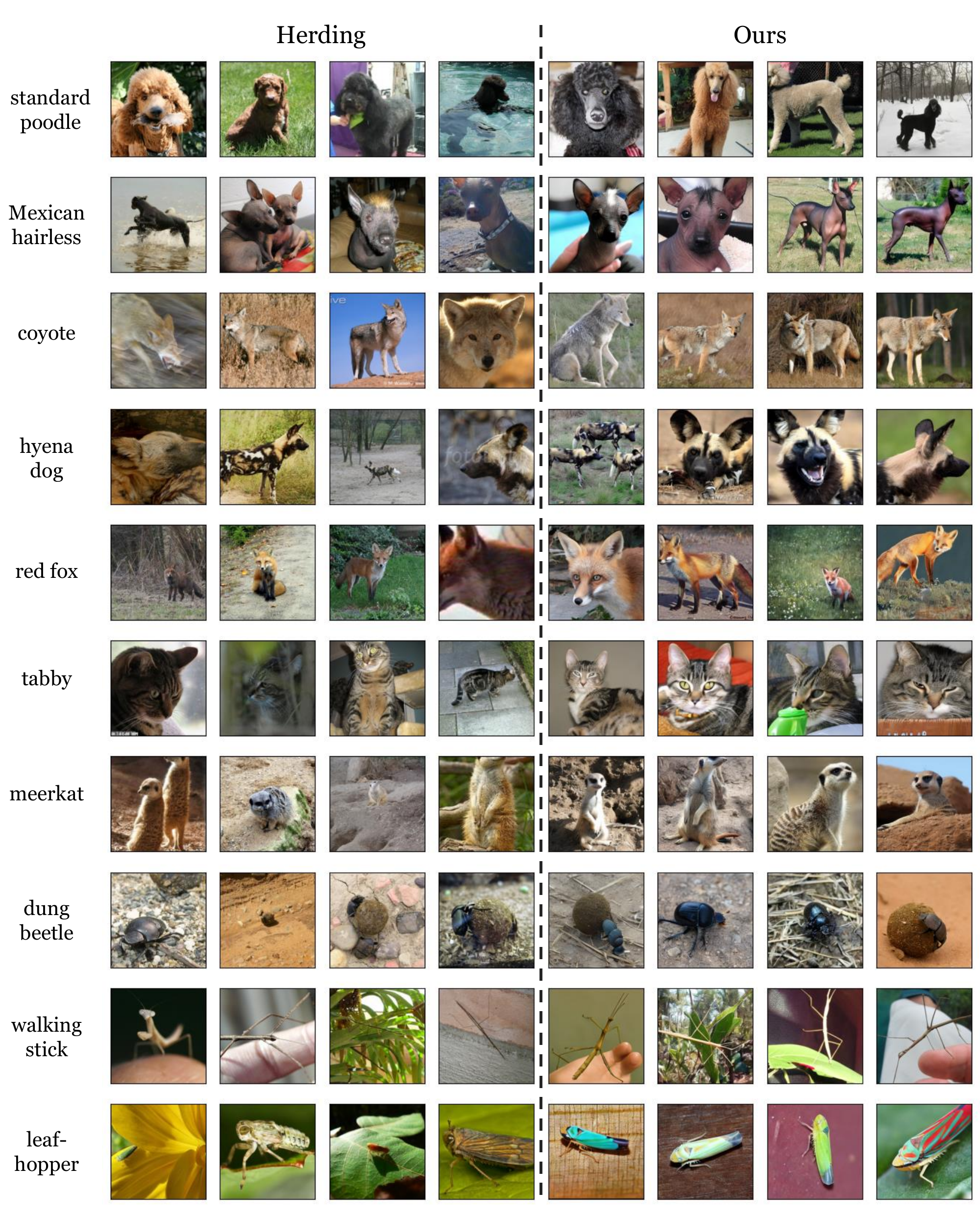}
    \caption{Comparison between samples selected by Herding (left) and generated by the proposed minimax diffusion method (right) for ImageNet-100 classes 30-39. The class names are marked at the left of each row. }
    \label{fig:phase3}
\end{figure*}

\begin{figure*}
    \centering
    \includegraphics[width=0.98\textwidth]{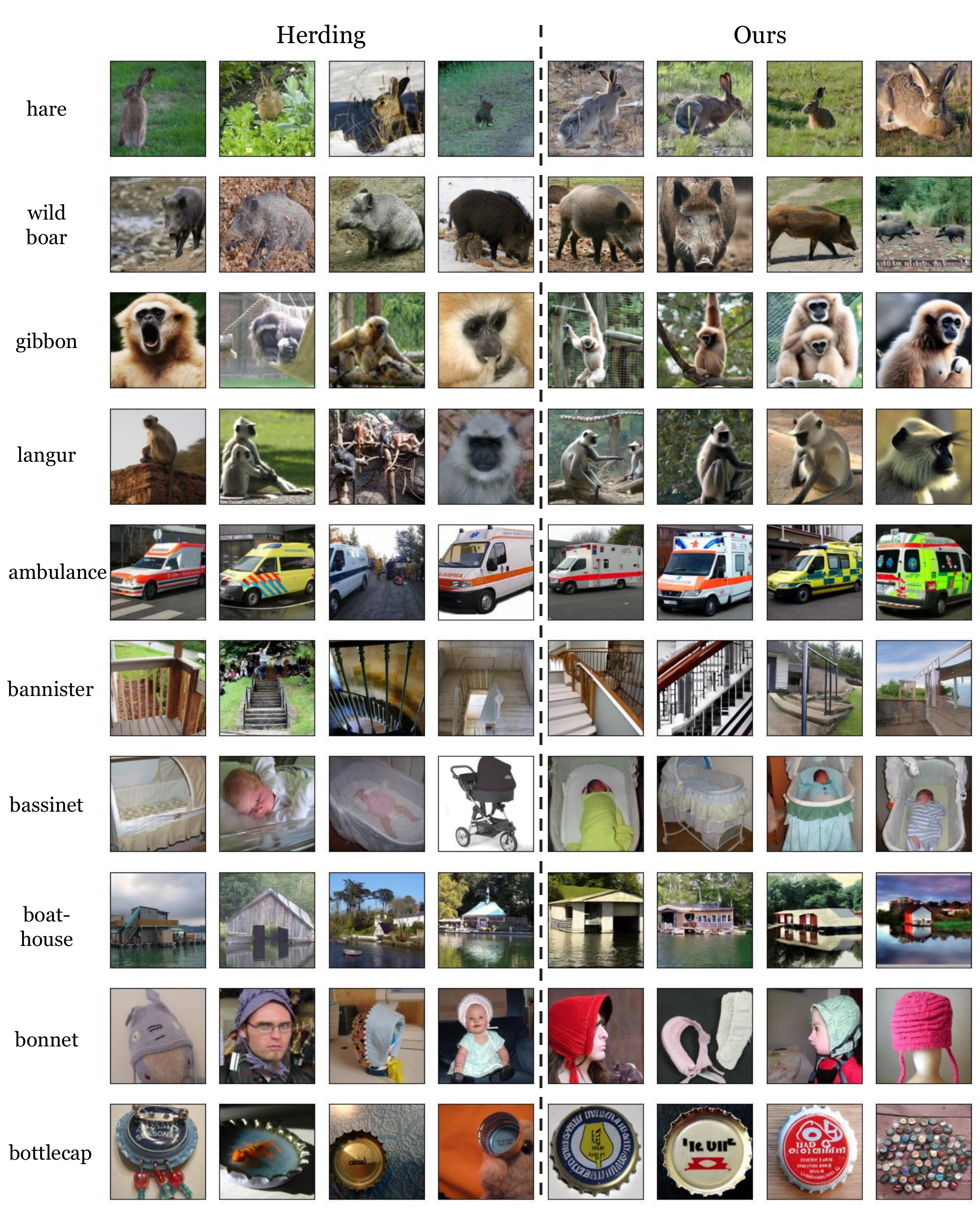}
    \caption{Comparison between samples selected by Herding (left) and generated by the proposed minimax diffusion method (right) for ImageNet-100 classes 40-49. The class names are marked at the left of each row. }
    \label{fig:phase4}
\end{figure*}

\begin{figure*}
    \centering
    \includegraphics[width=0.98\textwidth]{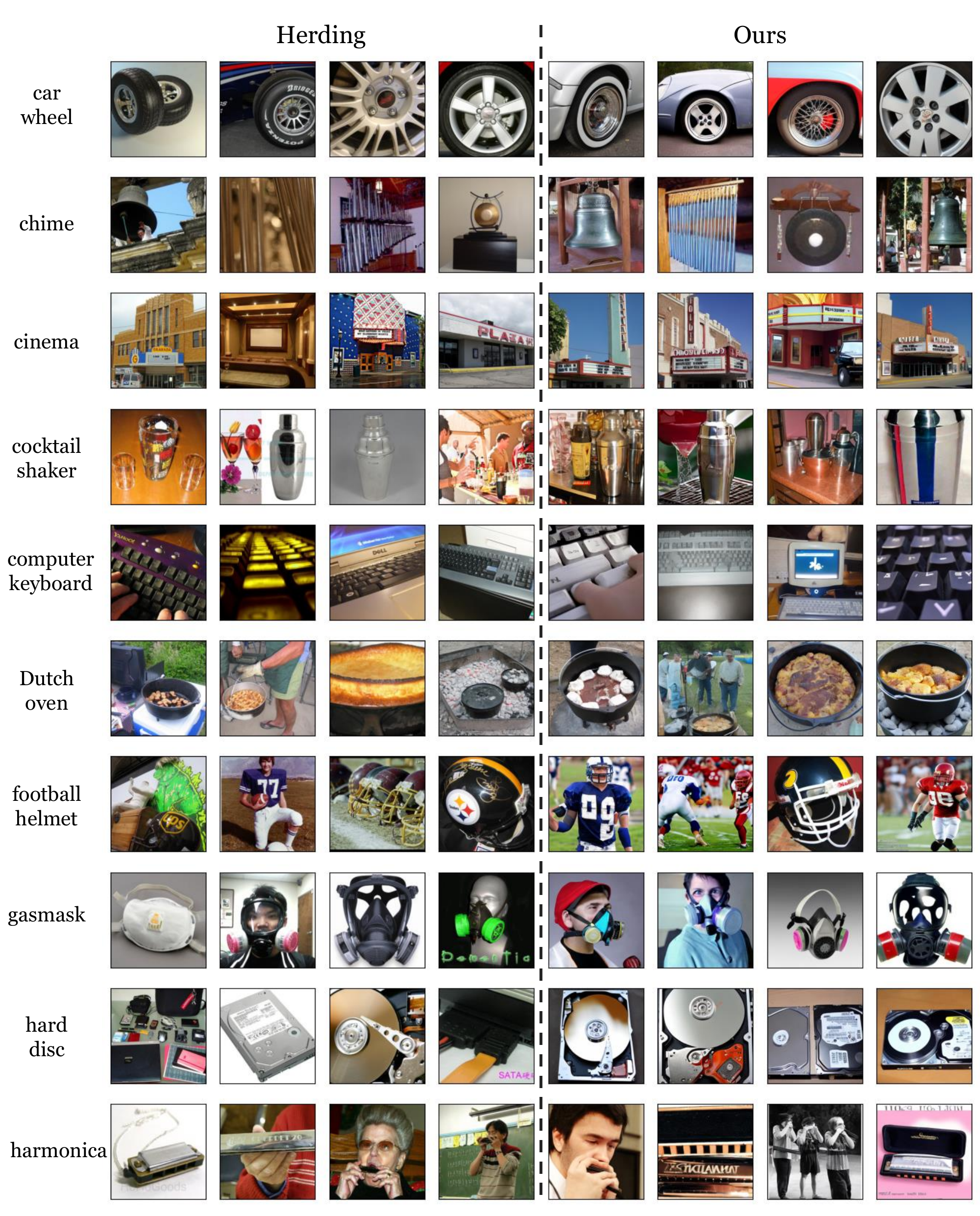}
    \caption{Comparison between samples selected by Herding (left) and generated by the proposed minimax diffusion method (right) for ImageNet-100 classes 50-59. The class names are marked at the left of each row. }
    \label{fig:phase5}
\end{figure*}

\begin{figure*}
    \centering
    \includegraphics[width=0.98\textwidth]{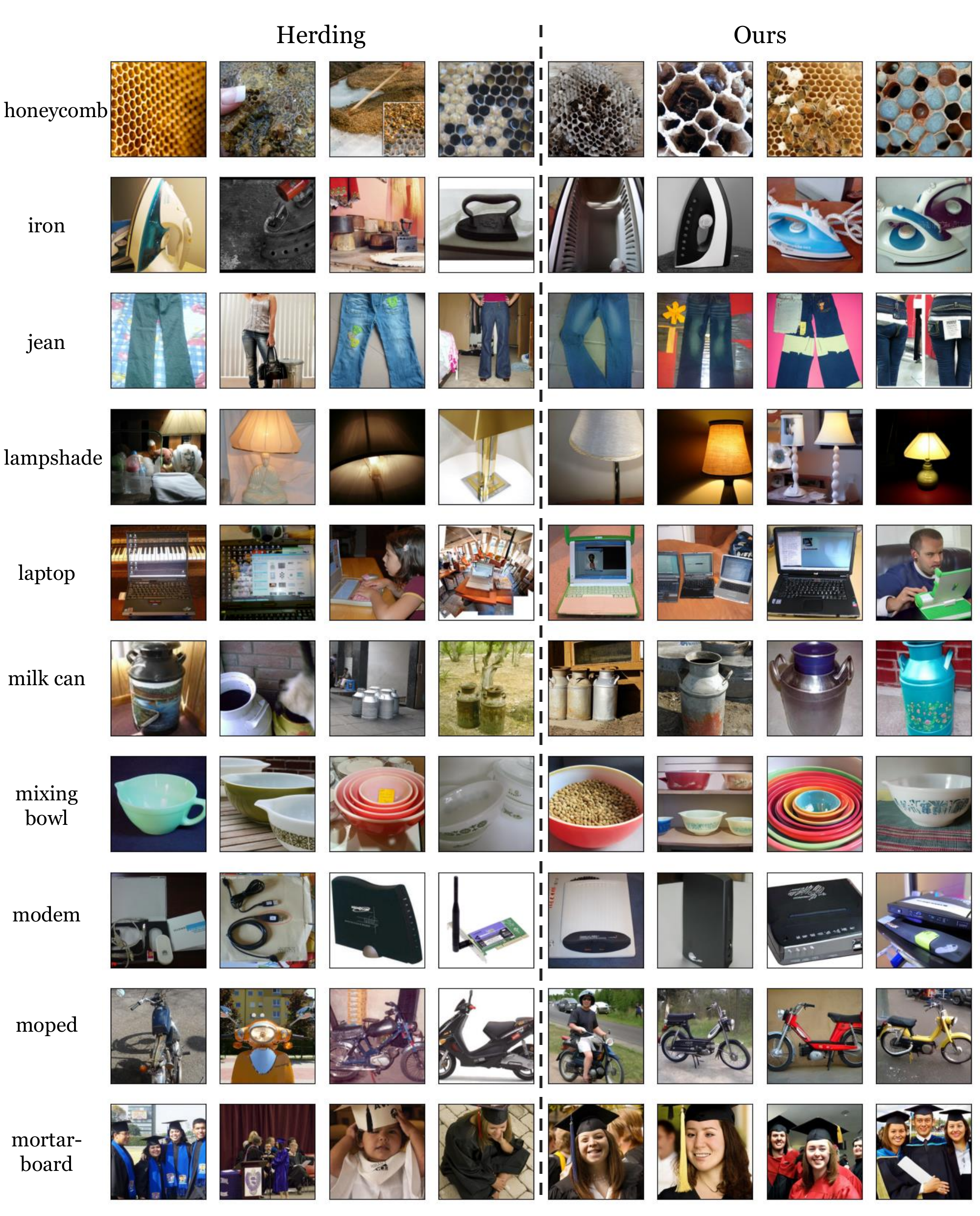}
    \caption{Comparison between samples selected by Herding (left) and generated by the proposed minimax diffusion method (right) for ImageNet-100 classes 60-69. The class names are marked at the left of each row. }
    \label{fig:phase6}
\end{figure*}

\begin{figure*}
    \centering
    \includegraphics[width=0.98\textwidth]{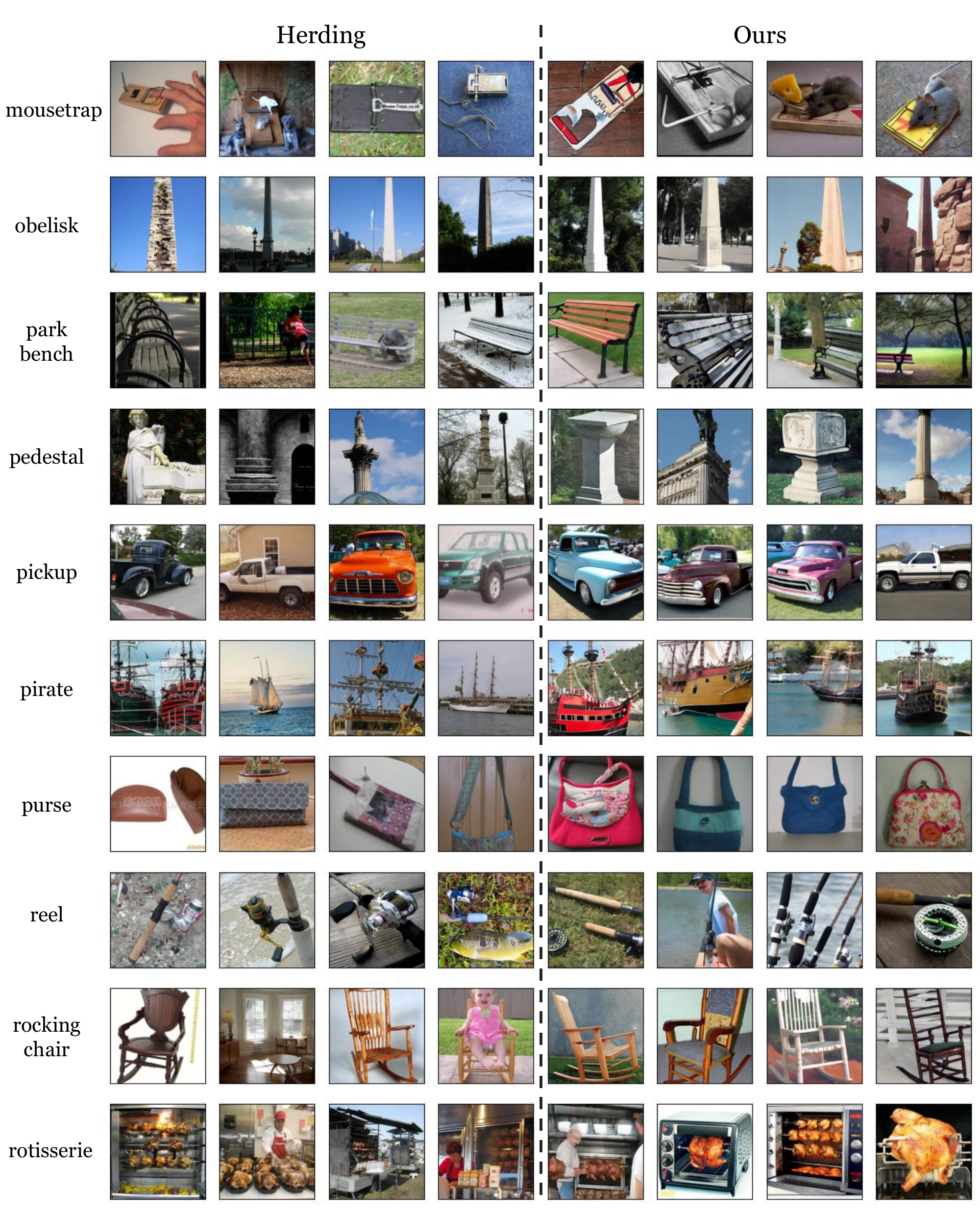}
    \caption{Comparison between samples selected by Herding (left) and generated by the proposed minimax diffusion method (right) for ImageNet-100 classes 70-79. The class names are marked at the left of each row. }
    \label{fig:phase7}
\end{figure*}

\begin{figure*}
    \centering
    \includegraphics[width=0.98\textwidth]{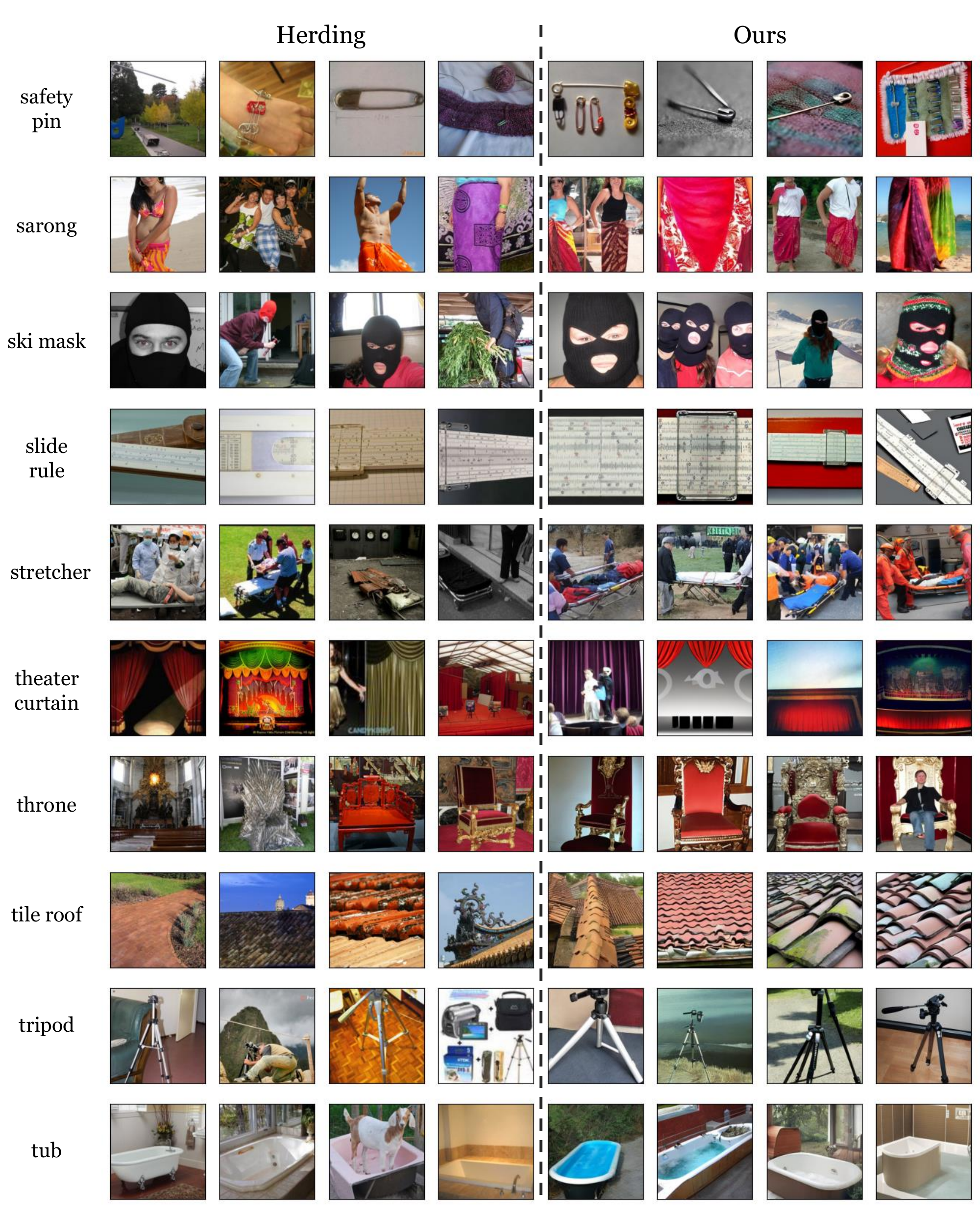}
    \caption{Comparison between samples selected by Herding (left) and generated by the proposed minimax diffusion method (right) for ImageNet-100 classes 80-89. The class names are marked at the left of each row. }
    \label{fig:phase8}
\end{figure*}

\begin{figure*}
    \centering
    \includegraphics[width=0.98\textwidth]{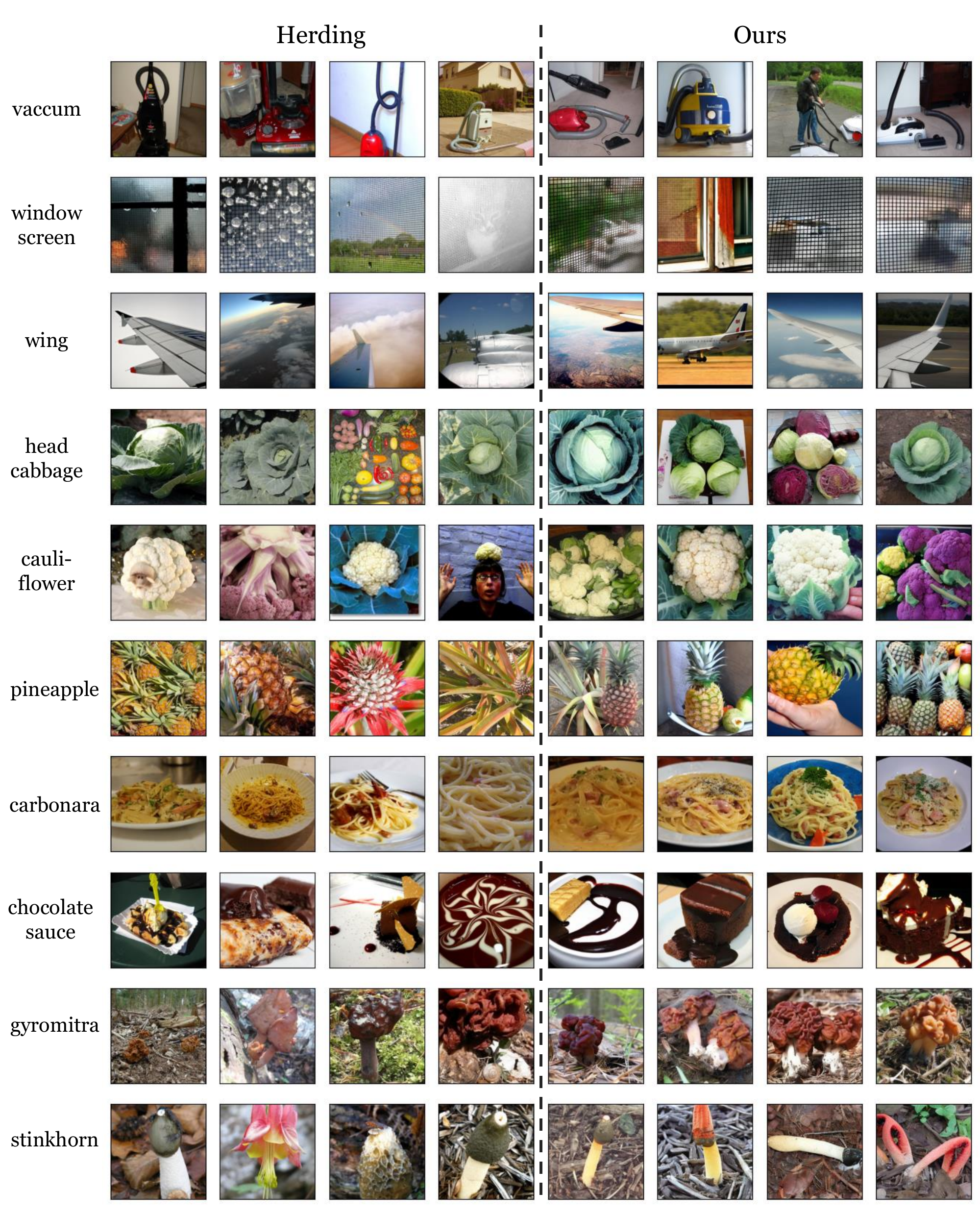}
    \caption{Comparison between samples selected by Herding (left) and generated by the proposed minimax diffusion method (right) for ImageNet-100 classes 90-99. The class names are marked at the left of each row. }
    \label{fig:phase9}
\end{figure*}

We present the comparison between the samples selected by Herding~\cite{welling2009herding} and those generated by our proposed minimax diffusion method on ImageNet-100 from~\cref{fig:phase0} to~\cref{fig:phase9}. 
In most cases, the diffusion model is able to generate realistic images, which cannot easily be told from real samples. 
Herding also aims to select both representative and diverse samples. 
However, the lack of supervision on the semantic level led to the inclusion of noisy samples. 
For instance, the walking stick class contains images of mantis, which can originally be caused by mislabeling. 
The proposed minimax diffusion, in comparison, accurately generates images of the corresponding classes, which is also validated by the better performance shown in~\cref{tab:imagenet-100}. 
There are also some failure cases for the diffusion model. 
The fur texture of hairy animals like Shih-Tzu and langur is unrealistic. 
The structures of human faces and hands also require further refinement. 
We treat these defects as exploration directions of future works for both diffusion models and the dataset distillation usage. 

\section{Broader Impacts} \label{broader-impact} 
The general purpose of dataset distillation is to reduce the demands of storage and computational resources for training deep neural networks. 
The requirement of saving resource consumption is even tenser at the age of foundation models. 
Dataset distillation aims to push forward the process of environmental contributions. 
From this perspective, the proposed minimax diffusion method significantly reduces the requirement resources for the distillation process itself. 
We hope that through this work, the computer vision society can put more attention on practical dataset distillation methods, which are able to promote the sustainable development of society. 

\section{Ethical Considerations} \label{Ethical}
There are no direct ethical issues attached to this work. 
We employ the publicly available ImageNet dataset for experiments. 
In future works, we will also be devoted to considering the generation bias and diversity during constructing a small surrogate dataset. 

\end{document}